\documentclass[preprint,10pt,3p]{elsarticle}
\usepackage[utf8]{inputenc}
\usepackage{algorithm}
\usepackage{algpseudocode}
\usepackage{amsmath}
\usepackage{amsfonts}
\usepackage{amssymb}
\usepackage{amsthm}
\usepackage{enumitem}
\usepackage{graphicx}
\usepackage{hyperref} 
\usepackage[dvipsnames]{xcolor}
\usepackage{tikz-cd}
\usepackage{tikz,pgfplots}
\usepackage{caption}
\usepackage{subcaption}
\usepackage{neuralnetwork}
\usepackage{cleveref}

\pgfplotsset{compat=1.16}

\newtheorem{definition}{Definition}

\newtheorem{assumption}{Assumption}

\newtheorem{remark}{Remark}
\newtheorem{example}{Example}

\Crefname{lemma}{Lemma}{Lemmas}
\crefname{section}{Section}{Sections}
\crefname{assumption}{Assumption}{Assumptions}
\crefname{proposition}{Proposition}{Propositions}
\crefname{corollary}{Corollary}{Corollaries}
\crefname{theorem}{Theorem}{Theorems}
\crefname{definition}{Definition}{Definitions}

\DeclareMathOperator{\rank}{rank}
\DeclareMathOperator{\Ker}{Ker}
\DeclareMathOperator{\Span}{Span}

\journal{}

\begin{document}
	
	\begin{frontmatter}
		
		
		
		\title{A singular Riemannian geometry approach to Deep Neural Networks II. Reconstruction of 1-D equivalence classes.}

		
		\author[ESP,GNCS]{Alessandro Benfenati\corref{cor1}}
		\ead{alessandro.benfenati@unimi.it}
		\ead{https://sites.unimi.it/a_benfenati/}
		\cortext[cor1]{Corresponding Author}
		
		\author[AM,GNFM,INFN]{Alessio Marta}
		\ead{alessio.marta@unimi.it}
		
		\affiliation[ESP]{organization={Environmental Science and Policy Department, Università di Milano},
			addressline={Via Celoria 2}, 
			city={Milano},
			postcode={20133}, 
			country={Italy}}
		\affiliation[GNCS]{organization={Gruppo Nazionale Calcolo Scientifico},
			addressline={INDAM},country={Italy}}

		\affiliation[AM]{organization={Dipartimento di Matematica, Universit{\`a} degli Studi di Milano},
			addressline={Via Cesare Saldini 50}, 
			city={Milan},
			postcode={20133}, 
			country={Italy}}
		\affiliation[GNFM]{organization={Gruppo Nazionale per la Fisica Matematica},
			addressline={INDAM},country={Italy}}            
		\affiliation[INFN]{organization={Istituto Nazionale di Fisica Nucleare, sezione di Milano},
			addressline={INFN},country={Italy}}            
		
		\begin{abstract}
			We proposed in a previous work a geometric framework to study a deep neural network, seen as sequence of maps between manifolds, employing singular Riemannian geometry. In this paper, we present an application of this framework, proposing a way to build the class of equivalence of an input point: such class is defined as the set of the points on the input manifold mapped to the same output by the neural network. In other words, we build the preimage of a point in the output manifold in the input space. In particular. we focus for simplicity on the case of neural networks maps from $n$--dimensional real spaces to ($n-1$)--dimensional real spaces, we propose an algorithm allowing to build the set of points lying on the same class of equivalence. This approach leads to two main applications: the generation of new synthetic data and it may provides some insights on how a classifier can be confused by small perturbation on the input data (\emph{e.g.} a penguin image classified as an image containing a chihuahua). In addition, for neural networks from 2D to 1D real spaces, we also discuss how to find the preimages of closed intervals of the real line. We also present some numerical experiments with several neural networks trained to perform non-linear regression tasks, including the case of a binary classifier.
		\end{abstract}

		\begin{highlights}
			\item The equivalence classes are sets of points that are mapped in the same way by the Neural Network
			
			\item We reconstruct equivalence classes of a given label in the NN input manifold 
			
			\item We develop a strategy to step from an equivalence class to another in input manifold
			
			\item We apply the developed algorithms to thermodynamics and classification problems
		\end{highlights}
		
		\begin{keyword}
			Deep Learning \sep Neural Network \sep Classification problem \sep Riemannian Geometry
			
			
			
		\end{keyword}
		
	\end{frontmatter}
	
	\section{Introduction}
	Neural Networks (NN) have been acknowledged to be a very powerful tool in machine learning tasks: A not fully comprehensive list of such tasks includes as speech--to--text transcription \cite{LIU2021306,Carlini18}, image segmentation \cite{Zhu20,GRIDACH2021274}, image classification \cite{CHUMACHENKO2022220}, match new items and/or products with user’s interests \cite{Lyu20}, image morphing \cite{eff21}, imitation learning \cite{SEKER202222}, solution to nonlinear PDEs \cite{RAISSI19,BLAKSETH2022181}, image generation \cite{Changhee18,SUN2020374}. The beginning of the new millennium has seen a growing interest in this automatic learning approach, due to the rising computational power (more performant GPUs) and the huge amount of data (the so-called Big Data Revolution). There exists several kind of neural networks, each one tailored for solving a particular problem: Some examples include Generative Adversarial Networks (GAN) \cite{Creswell18} for the generation of new images, U--Net for image segmentation \cite{Ibtehaz20}, Siamese Networks \cite{Iaroslav16} for image matching. The majority of the previous techniques handles data and NN's structures in an Euclidean framework: in this work we pursue our analysis by adopting a differential geometry approach.
	The idea that a neural network can be studied using differential geometry, working with non Euclidean data, is not new in literature \cite{HBL15,BBLSV17,HaRa17,M17,Shen18,GSGR19}. In particular in \cite{Shen18} and \cite{HaRa17} it is shown that a fully connected neural network can be studied as a sequence of maps between manifolds, the latter focusing on the case of Riemannian manifolds. In \cite{HaRa17} a notion of non-Euclidean distance between input points is proposed using a Riemannian metric tensor over the input manifold obtained employing the pullback of the Riemannian metric over the output manifold through the neural network map. However, the analysis carried out in \cite{HaRa17} focus on the case of a sequence of manifolds of the same dimension, since they assume that the pullback of the Riemannian metric through the neural network is still a Riemannian metric -- which is true only if the function realizing the neural network is a diffeomorphism: In general the resulting metric could be degenerate. To overcome this issue, we can assume to work in the setting of singular Riemannian geometry, in which the non-degeneracy of the metric is not assumed a priori. Starting from the ideas introduced in \cite{HaRa17}, in \cite{BeMa21} we introduced a geometric framework to study fully connected neural networks with singular Riemannian geometry and we proved several results about neural networks seen as a finite sequence of maps between manifolds of the form
	\begin{equation}
	\label{eq:sequence_of_maps}
	\begin{tikzcd}
	M_{0} \arrow[r, "\Lambda_1"] & M_{1} \arrow[r, "\Lambda_2"] & M_{2} \arrow[r,"\Lambda_3"]  & \cdots \arrow[r,"\Lambda_{n-1}"] & M_{n-1} \arrow[r, "\Lambda_n"] & M_n
	\end{tikzcd}
	\end{equation}
	with the maps $\Lambda_i$ realizing the layers of the network (see \cref{def:layer}) and $M_0, M_n$ being the input data manifold and the output manifold respectively. In the following we shall call the map $\mathcal{N}:=\Lambda_n \circ \cdots \circ \Lambda_1: M_0 \rightarrow M_n$ the neural network map. In this paper, for a matter of simplicity, we work with a particular class of manifolds, that is nonetheless sufficient for applications, assuming that every manifold of the sequence \eqref{eq:sequence_of_maps} is either $\mathbb{R}^{d_i}$ or a subset of $\mathbb{R}^{d_i}$. Note that this assumption is tantamount to require that every manifold admits a global coordinate system covering the whole set. Moreover, we consider a manifold $M=\mathbb{R}^n$ only as an affine space and not as a vector space. As an affine space, differences between two point - or displacement vectors - are well defined and correspond
	to tangent vectors over $M$. In particular, we can identify $T_p \mathbb{R}^n$ with $\mathbb{R}^n$ for every $p \in M$ \cite{Tu11}.
	
	Using this framework, we propose a way to build the class of equivalence of an input point, defined as the set of the points on the input manifold mapped to the same result by the neural network or, in other words, we build the preimage in $M_0$ of $\mathcal{N}(p)$ for a given point $p \in M_0$. Focusing, for simplicity, on the case of neural networks from subsets of $\mathbb{R}^n$ to subsets of $\mathbb{R}^{n-1}$, we propose an algorithm which allows to build a curve of points lying on the same class of equivalence and we present some numerical experiments in the case of a neural network trained to perform non-linear regression. The study of these equivalence classes may provide useful insights on the misbehaviour of some neural classifier: indeed, a small noise perturbation on the input data may lead the NN to completely misclassify even a simple image. For example, in \cite[Figure 1]{xie2018mitigating} a king penguin image is perturbed by a small amount of noise: such perturbation leads the classifier to label such noisy image as a chihuahua image. A further possible application is the generation of new synthetic data: indeed, the equivalence class constructed by the proposed procedure contains all the points classified with a particular label. Exploring this equivalence class may lead to find reliable and realistic new samples of the considered class.
	
	In addition we also implement -- in the case of a final manifold which is a subset of the real line -- an algorithm allowing to explore the input manifold changing equivalence class at each step, which in turn allow us to find the preimage $\mathcal{N}(I)$ for a given closed interval $I \subset \mathbb{R}$.
	
	The structure of the paper is the following. In \Cref{sec:geometric_preliminaries} we recall some notions of differential and singular Riemannian geometry, mainly to fix the notation we employ in this work. Then, following \cite{BeMa21} we introduce a geometric framework to analyse neural networks. In particular we consider a neural network as a finite sequence of maps between Riemannian manifolds satisfying certain properties stated in \Cref{sec:geometric_approach_to_nn} and we discuss the link between classes of equivalence of a neural networks and singular Riemannian geometry, using the results of \cite{BeMa21}. Then in \Cref{sec:algos} we introduce the one dimensional Singular Metric Equivalence Class algorithm (SiMEC) to build the class of equivalence of an input element. We introduce then the one dimensional Singular Metric Exploring algorithm (SiMExp) and we discuss how to employ both SiMEC and SiMExp to build the subset $\mathcal{S} \subset M_0$ such that $\mathcal{N}(\mathcal{S})=[a,b]$, $a,b \in \mathbb{R}$, or in other words, to find the preimage of $[a,b]$, a feature which is useful to study classifiers, since we can find all the points which are classified in the same way. \Cref{sec:numexp} is devoted to present some numerical experiments: the first set of experiments regard the SiMEC algorithm, where we apply our framework to a fully connected neural network trained to perform a nonlinear regression, learning some functions from $\mathbb{R}^2$ to $\mathbb{R}$, or equivalently learning some surfaces in $\mathbb{R}^3$ from a cloud of points. In the second part of this section we extend the previous tests to the SiMExp algorithm, including an application to binary classification, in which we build an approximate separating surface. The code employed to run these numerical experiments can be found at \url{http://github.com/alessiomarta/simec-1d-test-code}. 
	
	\paragraph{Notations} The set of positive real numbers is denoted with $\mathbb{R}_+$, while $\mathbb{R}_0^+$ denotes the set of non-negative real numbers. $T_p M$ is the tangent space of the smooth manifold $M$ over the point $p$ and $TM$ is the tangent bundle of the smooth manifold $M$. The space of the bilinear forms over the vector space $V$ valued in $\mathbb{R}_0^+$ is $Bil(V,V)$. $l(\gamma)$ stands for the length of a curve $\gamma$, while $Pl(\gamma)$ is the pseudolenght of a curve $\gamma$. Analogously, $d(x,y)$ is distance between the points $x$ and $y$ and $Pd(x,y)$ is pseudodistance between the points $x$ and $y$. $\dim(M)$ denotes the dimension of a manifold (or vector space) $M$. $\Ker(g)$ is the kernel of the matrix (or metric) $g$, while $\Span{V}$ is the set of all finite linear combinations of elements if $V$. Given a function $f:\mathbb{R}^n\to\mathbb{R}^m$, $J_f$ is its Jacobian matrix. $F^*$ denotes the pullback of the smooth map $F$.
	
	\section{Geometric preliminaries}
	\label{sec:geometric_preliminaries}
	
	\subsection{Basic notions in differential geometry}
	\label{subsec:basic_differential_geometry}
	In this section we briefly recall some basic notions in differential geometry we employ in the rest of the paper. The details can be found in any standard textbook about differential geometry, \emph{e.g.} \cite{Tu11} \cite{DoCarmo16}. See also \cite{BeMa21} for a more complete review of the background material. We begin reviewing the notion of smooth manifold, a generalization of the concept of surface in $\mathbb{R}^3$. We remember that a topological space $M$ is Hausdorff (or $T_2$) if for any two points $p,q \in M$, then there are two sets $S_p$ and $S_q$ such that $S_p \cap S_q = \emptyset$; A topological space is called second countable if it admits a countable basis.
	\begin{definition}
		A smooth d-dimensional manifold $M$ is a second countable and Hausdorff topological space such that every point $p \in M$ has a neighbourhood $U_p$ that is homeomorphic to $\mathbb{R}^d$ through a map $\phi_p:U_p \rightarrow \mathbb{R}^{d}$, with the additional requirement that if $U_p \cap U_q \neq \emptyset$, then $\phi_p \circ \phi_q^{-1}$ is a smooth diffeomorphism. 
	\end{definition}
	The pair $(U_p,\phi_p)$ is called a \emph{local chart} and the collection of all the possible local charts at all points is an \emph{atlas}. If we can cover the whole manifold $M$ with a chart, it will be called a \emph{global chart}.
	The definition given above is not the most general one, but it is sufficient for our purposes. Indeed, in the following we are interested in smooth manifolds of the form $M=(a_1,b_1) \times \cdots \times(a_n,b_n) \subset \mathbb{R}^n$, $a_i < b_i \ \forall i=1,\cdots,n$. Any subset of the Euclidean space with this form is diffeomorphic to $\mathbb{R}^n$ and admits a global chart inherited from $\mathbb{R}^n$, for example the Cartesian coordinates restricted to $(a_1,b_1) \times \cdots (a_n,b_n)$. Given a function $F$ between two smooth manifold $M$ and $N$, we say that $F$ is smooth if for any choice of local charts $(U_M,\phi_M),(U_N,\phi_N)$ of $M$ and $N$ respectively, the function $\phi_N \circ F \circ \phi_M^{-1} : U_M \subset \mathbb{R}^{\dim(M)} \rightarrow U_N \subset \mathbb{R}^{\dim(N)}$ is smooth in the usual sense. In the case $F$ is a function between two subsets of two Euclidean spaces $\mathbb{R}^{l}$ and $\mathbb{R}^m$, $l,m \in \mathbb{N}$, this definition is tantamount to require that the components of the map $F$ are smooth. A notable example of function over manifolds is a curve over a manifold $M$, namely a smooth function $\gamma : \mathbb{R} \rightarrow M$, $t \mapsto \gamma(t)$.
	
	In practice, describing a manifold giving an atlas or (when possible) a global chart is not always simple. A common (and simpler) way to build a manifold is to embed it in a higher dimensional space. For example we can realize the unit sphere $S^2$ -- a 2-dimensional manifold -- as the subset of $\mathbb{R}^3$ -- a 3-dimensional manifold with the usual Cartesian coordinates $(x,y,z)$ -- such that $x^2+y^2+z^2=1$ and  we say that we embedded $S^2$ in $\mathbb{R}^3$. If we describe a manifold $M$ as a subset of the Euclidean space $\mathbb{R}^n$, we say that $M$ is embedded in $\mathbb{R}^n$, whose in turn is called embedding space. The map realizing $M$ as a subspace of $\mathbb{R}^n$ is called embedding. In general the definition of embedding from a manifold $M$ to a manifold $N$ is the following.
	\begin{definition}\label{def:embedding}
		Let $f:M \rightarrow N$ be a smooth map between manifolds. $f$ is an embedding if its differential is everywhere injective and if it is an homeomorphism with its image. In other words, $f$ is a diffeomorphism with its image.
	\end{definition}
	The next notion we need to recall is that of tangent space, a generalization of the tangent plane to a two-dimensional surface. For our purposes, it is convenient to define this geometric object using tangent vectors to curves over $M$. Let $p \in M$ and consider a chart $(U,\phi)$ containing $p$. Consider two smooth curves $\gamma_1,\gamma_2:(-1,1) \rightarrow U$ such that $\gamma_1(0)=\gamma_2(0)=p$. We say that $\gamma_1$ is equivalent to $\gamma_2$ at $p$ if $\frac{d}{dt}\left(\phi \circ \gamma_1\right)(t)|_{t=0}=\frac{d}{dr}\left(\phi \circ \gamma_2\right)(t)|_{t=0}$ (see \cite[Figure 1]{BeMa21}). 
	\begin{definition}\label{def:tangent_vectors}
		A tangent vector $\gamma^\prime(0)$ over $p$ is an equivalence classes of curves and the tangent space of $M$ at $p$, which is denoted with $T_p M$, is the set of all tangent vectors at $p$.
	\end{definition}
	
	The definition of tangent vector is not dependent on a particular choice of a chart. 
	Taking the disjoint union of tangent spaces at all points, we obtain the tangent bundle $TM$. A choice of a basis for $T_pM$ is called reference frame.
	\begin{example}
		Let $M=\mathbb{R}^n$ considered as an affine space or $M=(a_1,b_1) \times \cdots  \times(a_n,b_n) \subset \mathbb{R}^n$. $M$ is a smooth manifold whose tangent space $T_p M$ over each point coincides with the vector space $\mathbb{R}^n$, namely the space of the displacement vectors. The tangent bundle is given by $TM = M \times \mathbb{R}^n$. 
	\end{example}
	A smooth function $F:M \rightarrow N$ between two smooth manifolds $M$ and $N$ is also mapping vectors in $TM$ to vectors in $TN$. An interesting class of maps between manifold sending the whole $TM$ to $TN$ is that of submersions.
	\begin{definition}\label{def:submersion}
		Let $f:M \rightarrow N$ be a smooth map between manifolds. Then $f$ is a submersion if, in any chart, the Jacobian $J_f$ has rank $\dim(N)$.
	\end{definition}
	
	\subsection{Riemannian geometry}\label{subsec:riemannian_geometry}
	We continue this brief excursus on differential geometry by recalling some basic notions of Riemann geometry, focusing on the case of manifolds realized as subsets of $\mathbb{R}^n$. The pivotal notion in Riemannian geometry is that of Riemannian metric, defined on a generic manifold as follows.
	\begin{definition}\label{def:general_metric}
		A Riemannian metric $g$ over a smooth manifold $M$ is a smooth family of inner products on the tangent spaces of $M$; Namely, $g$ associates to every $p \in M$ an inner product $g_p: T_p M \times T_p M \rightarrow \mathbb{R}$, with $g_p(x,y)=0$ if and only of $x=0$ or $y=0$. 
	\end{definition}
	Given a metric $g$, we define the norm of a vector $v \in T_p M$ as $\| v \|_p = \sqrt{g_p(v,v)}$. When a manifold is either $\mathbb{R}^n$ or a subset $I\subset \mathbb{R}^n$, $I=(a_1,b_1)\times \dots\times(a_n,b_n)$, of it w we can identify $T_p \mathbb{R}^n$ with $\mathbb{R}^n$ itself \cite{Tu11}. In particular we can simplify \Cref{def:general_metric} to the following one.
	\begin{definition}\label{def:simplified_metric}
		Consider a smooth manifold $M$, $M=\mathbb{R}^n$ or $M=(a_1,b_1)\times \dots\times(a_n,b_n)\subset\mathbb{R}^n$. A Riemannian metric $g$ over $M$ is a map $g:M \rightarrow Bil(\mathbb{R}^n \times \mathbb{R}^n)$ that associates to each point $x$ a positive symmetric bilinear form $g_x:\mathbb{R}^n \times \mathbb{R}^n \rightarrow \mathbb{R}$ in a smooth way.
	\end{definition}
	\begin{remark}
		Even if we can specialize definition \eqref{def:general_metric} to \eqref{def:simplified_metric}, in accordance to \cite{Tu11,DoCarmo16}, it is important not to confuse $M$ and its tangent space. Since one should think of $M=\mathbb{R}^n$ as an affine space and of $T_p M \simeq \mathbb{R}^n$ as the space of displacement vectors, a metric $g$ associates to every point $x \in M$ a bilinear form over displacement vectors.
	\end{remark}
	Fully specifying a metric is enough to know the matrix associated to $g$ in a coordinate system $(x_1,\cdots,x_n)$. Let $(y_1,\cdots,y_n)$ be another coordinate system related to the original one by the diffeomorphism $\varphi:M \rightarrow M$, $(y_1,\cdots,y_n) \mapsto (x_1,\cdots,x_n)$. Then the matrix representing the metric $g$ in the new coordinates is given by
	\begin{equation}\label{eq:metric_change_of_charts} 
	g^y_{ij} =  \sum_{h,k}\left( \frac{\partial x^h}{\partial y^i} \right) g^x_{hk} \left( \frac{\partial x^k}{\partial y^j} \right)
	\end{equation}
	where $\displaystyle\frac{\partial x^h}{\partial y^i} = (J_\varphi)_{hi}$ is the $(h,i)$ entry of the Jacobian of the function $\varphi$, $g^x$ and $g^y$ denotes the metric in the $x$--coordinates and in the $y$--coordinates systems, respectively. If a manifold $M$ can be realized as a subset of the Euclidean space, it naturally inherits a canonical Riemannian metric from $\mathbb{R}^n$. Let $	\eta : M \rightarrow \mathbb{R}^n$ be the immersion map of $M$ in $\mathbb{R}^n$. Then we can equip $M$ with the metric
	\begin{equation}\label{eq:induced_metric_rn} 
	h = J_\eta^T g J_\eta
	\end{equation}
	where $J_\eta$ is the Jacobian matrix of $\eta$ in local coordinates, namely $(J_\eta)_{ij} = \left( \dfrac{\partial \eta^i}{\partial x^j} \right)$. If a manifold $M$ is equipped with a Riemannian metric $g$, we have a canonical definition of length of a curve. 
	\begin{definition}\label{def:lenght}
		Let $\gamma:[a,b] \rightarrow M$ be a piecewise smooth curve, then its length is 
		\begin{equation}
		L(\gamma) = \int_a^b \| \dot{\gamma}(s) \|_{\gamma(s)} ds = \int_a^b \sqrt{g_{\gamma(s)}(\dot{\gamma}(s),\dot{\gamma}(s))} ds
		\end{equation}
	\end{definition}
	Notice that, given a curve $\gamma:[a,b] \rightarrow M$, $L(\gamma)=0$ if and only if $\gamma$ is the constant map associating to every $s \in [a,b]$ the same point $p\in M$. This is a consequence of the non-degeneracy of the Riemannian metric. If a manifold $M$ is path-connected we can also define the distance function $d:M \times M \rightarrow \mathbb{R}_0^+$ as follows. 
	\begin{equation*}
	\begin{split}
	d(x,y) = \inf \{ l(\gamma) \ | \ \gamma:[0,1]\rightarrow M \mbox{ is a piecewise } \mathcal{C}^1 \mbox{ curve with } \gamma(0)=x \mbox{ and } \gamma(1)=y \}   
	\end{split}
	\end{equation*}
	for every $x,y \in M$. As a result, the pair $(M,d)$ is a metric space.
	Another notion closely related to the length of a curve is that of energy of a curve, defined for a piecewise $\mathcal{C}^1$ curve $\gamma:[0,1] \rightarrow M$ as
	\begin{equation}
	E(\gamma) = \int_0^1  \| \dot{\gamma}(s) \|^2_{\gamma(s)} ds =  \int_a^b g_{\gamma(s)}(\dot{\gamma}(s),\dot{\gamma}(s)) ds
	\end{equation}
	In Riemannian geometry $l(\gamma)= 0$ if and only if $E(\gamma)=0$ and, in general, a curve minimizes the length functional if and only if it minimizes the energy functional.
	Consider two smooth manifolds $M,N$ and let $F:M \rightarrow N$ be a smooth map. Suppose that $N$ is equipped with a Riemannian metric $g^N$. Then we can endow $M$ with the pullback metric $g^M = F^* g^N$. Chosen two global coordinate systems $(x_1,\cdots,x_m)$ and $(y_1,\cdots,y_n)$ of $M$ and $N$ respectively, the matrix associated to the pullback of $g^N$ through $F$ reads:
	\begin{equation}\label{eq:metric_pullback}
	(g^M)_{ij} = \sum_{h,k = 1}^{\dim(N)} \left( \frac{\partial F^h}{\partial x^i} \right) (g^N)_{hk} \left( \frac{\partial F^k}{\partial x^j} \right)
	\end{equation}
	
	\Cref{eq:metric_change_of_charts} and \Cref{eq:induced_metric_rn} are particular cases of \Cref{eq:metric_pullback}. Moreover, 
	\Cref{eq:metric_pullback} allows to compute the Riemannian metric of manifolds realized as subsets of other Riemannian manifolds.
	When $F:M \rightarrow N$ is a diffeomorphism or $F$ is an immersion, encompassing the case in which $\dim(M)<\dim(N)$ and $J_F$ is injective, then the metric $g^M = F^* g^N$ obtained using the pullback of $g^N$ trough $F$ is still non-degenerate; However, if $J_F$ is not injective, and this is certainly true if $\dim(M)>\dim(N)$, then $g^M = F^* g^N$ is degenerate - the matrix associated with $g^M$ is not of full rank - and therefore it is not a Riemannian metric.
	
	\subsection{Singular Riemannian geometry}
	\label{subsec:singular_riemannian_geometry}
	In this section we introduce some notions of singular Riemannian geometry, which will be useful in the following. 
	\begin{definition}\label{def:general_singular_metric}
		A singular Riemannian metric $g$ over a smooth manifold $M$ is a smooth family of positive semi definite symmetric bilinear forms on the tangent spaces of $M$. 
	\end{definition}
	Given a singular metric $g$, we define the semi--norm of a vector $v \in T_p M$ as $\| v \|_p = \sqrt{g_p(v,v)}$. As we did for the definition of Riemannian manifold, we can specialize the definition of singular Riemannian metric to a particular class of manifolds, used in the forthcoming.
	\begin{definition}\label{def:simplified_singular_metric}
		Consider a smooth manifold $M$, $M=\mathbb{R}^n$ or $M=(a_1,b_1)\times \dots\times(a_n,b_n)\subset\mathbb{R}^n$.
		A singular Riemannian metric $g$ over $M$ is a map $g:M \rightarrow Bil(\mathbb{R}^n \times \mathbb{R}^n)$ that associates to each point $p$ a positive semidefinite symmetric bilinear form $g_p:\mathbb{R}^n \times \mathbb{R}^n \rightarrow \mathbb{R}$ in a smooth way.
	\end{definition}
	In singular Riemannian geometry, a singular metric $g_p(x,y)$ may be equal to 0 even if both $x\neq 0$ and $y\neq0$. All the formulae stated in the previous sections still hold true, even in the case of a singular Riemannian metric.  The sole difference is that we may have smooth non-constant curves, \emph{i.e.} those curves whose image is a single point, of null length. The degeneracy of the metric induces a decomposition of $T_p M$ over every point $p \in M$ as the direct sum $T_p P \oplus T_p N$, with $T_p P$ the subspace of vectors with positive seminorm and $T_p N$ the subspace spanned by the vectors whose seminorm is zero. With abuse of notation, we define the pseudolenght $Pl$ as
	\begin{equation}
	Pl(\gamma) = \int_a^b \| \dot{\gamma}(s) \|_{\gamma(s)} ds = \int_a^b \sqrt{g_{\gamma(s)}(\dot{\gamma}(s),\dot{\gamma}(s))} ds
	\end{equation}
	and the pseudodistance $Pd$
	\begin{equation}
	Pd(x,y) = \inf \{ l(\gamma) \ | \ \gamma:[0,1]\rightarrow M \mbox{ is a piecewise } \mathcal{C}^1 \mbox{ curve with } \gamma(0)=x \mbox{ and } \gamma(1)=y \}  
	\end{equation}
	and the previous observation about the existence of non-trivial curves of length zero in $M$ entails that there are points whose distance is null or, in other words, the pair $(M,Pd)$ is a pseudometric space. Identifying the metrically indistinguishable points using the equivalence relation $x \sim y \Leftrightarrow Pd(x,y)=0$ for $x,y \in M$, we obtain the metric space $(M_i / \sim,Pd)$. Note that a class of equivalence $[x]$ is the set $\{y \in M \ | \ Pd(x,y) = 0\}$ and therefore the points of $M/\sim$ are classes of equivalence of points in $M$. A curve whose pseudolength is zero is called null curve. See \cite[Example 5]{BeMa21} for a detailed insight on this aspects and the subsequent \Cref{ex:pullback_layer,ex:example_metric_ODE} for the case of nonlinear functions.
	
	\section{A geometric approach to neural networks} 
	\label{sec:geometric_approach_to_nn}
	
	\subsection{The geometric framework}
	\label{subsec:geometric_framework}
	In this section, we give a definition of neural networks as a finite sequence of maps between manifolds. In addition to the assumptions done in \cite{BeMa21}, we also make the following preliminary hypothesis, which was already used in the previous sections.
	\begin{assumption}\label{as_1}
		Every manifold of the sequence \eqref{eq:sequence_of_maps} is either $\mathbb{R}^{d_i}$ or a subset of $\mathbb{R}^{d_i}$ of the form $(a_1,b_1)\times\cdots\times(a_{d_i},b_{d_i})$. 
	\end{assumption}
	The rationale behind this hypothesis is that in practice we usually realize low-dimensional non-Euclidean manifolds using embeddings in higher-dimensional Euclidean spaces. Indeed, often it happens that we do not even know the non-Euclidean structure underlying a cloud of points contained in a dataset.
	
	As in \cite{BeMa21} we assume that all maps $\Lambda_i$ in the \eqref{eq:sequence_of_maps} are either embedding or submersion. 
	\begin{assumption}\label{as_2}
		The sequence of maps \eqref{eq:sequence_of_maps} satisfies the following properties:
		\begin{enumerate}[label=\arabic*)]
			\item If $\dim(M_{i-1}) \leq \dim(M_i)$ the map $\Lambda_i: M_{i-1} \rightarrow M_i$ is a smooth embedding.
			\item If $\dim(M_{i-1}) > \dim(M_i)$ the map $\Lambda_i: M_{i-1} \rightarrow M_i$ is a smooth submersion.
		\end{enumerate}
	\end{assumption}
	We provide the definition of smooth feedforward layer as a particular kind of maps between manifold.
	\begin{definition}[Smooth layer]\label{def:layer}
		Let $M_{i-1}$ and $M_i$ be two smooth manifolds in sequence \eqref{eq:sequence_of_maps} abiding to \Cref{as_1}. A map $\Lambda_i : M_{i-1} \rightarrow M_i = \Lambda_i(M_{i-1})$ is called a smooth layer if it is the restriction to $M_{i-1}$ of a function $\overline{\Lambda}_i(x)  : \mathbb{R}^{d_{i-1}} \rightarrow \mathbb{R}^{d_i}$ of the form
		\begin{equation}
		\overline{\Lambda}_i^\alpha(x) = F_i^\alpha\left(\sum_\beta A^{\alpha\beta}_i x_\beta+b_i^\alpha\right) 
		\end{equation}
		for $i=1,\cdots,n$, $\alpha=1,\cdots, d_{i-1}$, $x \in \mathbb{R}^{d_{i}}$, $b \in \mathbb{R}^{d_i}$ and $A \in \mathbb{R}^{d_{i} \times d_{i-1}}$, with $F_i: \mathbb{R}^{d_i} \rightarrow \mathbb{R}^{d_i} $ a diffeomorphism.
	\end{definition}
	The matrices $A_i$ and the vectors $b_i$ are the matrices of weights and the biases of the $i-th$ layer, while the functions $F_i$ are the activation functions.
	\begin{example}
		Classical activation functions such as sigmoid, softmax and softplus \cite{Agg18,GBC16,DFO20} satisfies \Cref{def:layer}. An example of commonly employed activation function not encompassed by our definition of layer is ReLu. In a ReLu layer $f :\mathbb{R}^{l} \to \mathbb{R}^{l}$,  $F^j(x)= \max\{0,x_j\}$ is not even $\mathcal{C}^1$. However we note that softplus layers are used as smooth approximations of ReLu layers. Moreover, employing smooth activation functions may lead to remarkable results in Deep Learning problems (see the SIREN architecture \cite{sitzmann2020implicit} for Signal Processing problems, which employs $\sin$ activation functions, or the IKE-XAI architecture \cite{CHRAIBIKAADOUD202295} using sigmoid and hyperbolic tangent functions for the knowledge construction of an artificial agent).
	\end{example}
	In order to satisfy \Cref{as_2}, we assume the following hypothesis.
	\begin{assumption}[Full rank hypothesis]
		We assume that the matrices of weights are of full rank.
	\end{assumption}
	This may seem a difficult requirement to met, even if it is quite common in literature \cite{Shen18,10.5555/3305890.3305950,Shen_2018_CVPR}, but in neural network training the set of matrices which are not full rank is a set of null measure: see \cite[remark 8]{BeMa21} for a thorough discussion.
	\begin{definition}[Smooth Neural Network]
		A smooth neural network is a sequence of maps between manifolds 
		\begin{equation}
		\label{eq:smoothNN}
		\begin{tikzcd}
		M_{0} \arrow[r, "\Lambda_1"] & M_{1} \arrow[r, "\Lambda_2"] & M_{2} \arrow[r,"\Lambda_3"]  & \cdots \arrow[r,"\Lambda_{n-1}"] & M_{n-1} \arrow[r, "\Lambda_n"] & M_n
		\end{tikzcd}
		\end{equation}
		with $n \geq 2$ such that the manifolds $M_i$ abide to \Cref{as_1} and the maps $\Lambda_i$ are as in \Cref{def:layer}. A Neural Network with only one hidden layer ($n=2$) is named \emph{shallow network}, whilst a neural network with more than two layers, namely with $n \geq 3$, is called a \emph{deep neural network}.
		\label{def:smoothNN}
	\end{definition}
	We will call $M_0$ the input data manifold, $M_n$ the output manifold and the other manifolds $M_i$ the representation manifolds. The map $\mathcal{N}=\Lambda_1\circ \cdots \circ\Lambda_n$ is the neural network map.

	\subsection{A singular Riemannian approach to Neural Networks}
	\label{subsec:singular_riemannian_approach_to_nn}
	Suppose now to endow the output manifold $M_n$ with a Riemannian metric $g^n$, for example the Euclidean metric for a regression task. Then we can equip the other manifolds $M_i$ of the sequence \eqref{eq:smoothNN} with a (in general singular) Riemannian metric $g^i$ via the pullback of $g^n$ through the layers $\Lambda_i$, namely $g^i = \Lambda_i^* \cdots \Lambda_n^* g^n$. We focus on $g^0 = \mathcal{N}^* g^n$ for simplicity, but one can repeat the same reasoning for all the other metrics $g^i$. From \cite{BeMa21} we know that, assuming the full rank hypothesis, $rank(g^0) \leq \min \{\dim(M_1),\cdots,\dim(M_n)\}$. 
	\begin{remark}\label{rem:rank_no_bottlenecks}
		In particular, in the case of a neural network without bottlenecks we have that $\rank(g^0) = \dim(\mathcal{N}(M_0))$, a fact which we may know a priori. This happens in some important cases: 
		\begin{itemize}
			\item For a neural network trained to perform non-linear regression from $M_0 \subset \mathbb{R}^{d_0}$ to $M_n \subset \mathbb{R}^{d_n}$, the rank of $g^0$ is $\dim(M_n)=d_n$.
			\item For a classifier whose last layer is of dimension $d_n$, the rank of the metric $g^0$ is $d_n-1$. Indeed, the output manifold $M_n$ is of dimension $d_n$, but we have one constraint: The sum of the output of the last layer must be $1$. Therefore $\mathcal{N}(M_0) \subset M_n$ is of dimension $d_n-1$.
		\end{itemize}
	\end{remark}
	We note that the analysis carried out in \cite{BeMa21} yields that $\dim(\Ker(g^0_p))$ does not depend on $p \in M_0$. In addition, if $\dim(M_0) \leq \dim(M_i)$ for every $i=1,\cdots,n$, then $g^0$ is a Riemannian metric; Otherwise $g^0$ is degenerate, and therefore a singular Riemannian metric.
	
	A key property is that \cite[Proposition 3]{BeMa21} two point $x,y \in M_0$ in the same class of equivalence are such that $\mathcal{N}(x)=\mathcal{N}(y)$. Calling $k = \mathcal{N}(x)$, this is tantamount to say that $y \in \mathcal{N}^{-1}(k)$ such that $x$ and $y$ are connected by a null curve. If $x \sim y$, by definition of $\sim$, there is a $\mathcal{C}^1$ curve $\gamma:(0,1)  \rightarrow M_0$ such that $\gamma(0)=x$ and $\gamma(1)=y$ with $\dot{\gamma}(s) \in \Ker(g^0_{\gamma(s)})$. On the other hand, given a null curve $\gamma:(0,1)\rightarrow M_0$, then all the points $\gamma(s)$, $s \in (0,1)$, belong to the same class of equivalence. Furthermore \cite[Remark 11]{BeMa21} there is a smooth basis of $\Ker(g_x)$ depending smoothly on $x$.
	This observation yields an algorithm allowing one to build a class of equivalence, topic of the next section. Before to discuss the SiMEC algorithm, we present two examples.
	
	\begin{example}
		\label{ex:pullback_layer}
		Consider the shallow neural network
		$$
		\begin{tikzcd}
		M_0 = \mathbb{R}^3 \arrow[r, "\Lambda_1"] & M_1 = \mathbb{R}^2 \arrow[r, "\Lambda_2"] & M_2 = \mathbb{R}^2
		\end{tikzcd}
		$$
		with $\Lambda_1$ and $\Lambda_2$ two sigmoid layers. Notice that $\Lambda_1(M_0) = (0,1) \times (0,1) \subset M_1$ and $\mathcal{N}(M_0) = \Lambda_2(\Lambda_1(M_0))  \subset (0,1) \times (0,1)$. Suppose to endow $M_2$ with a Riemannian metric $g^2$. \Cref{eq:metric_pullback} yields that $g^0=\Lambda^*_1(\Lambda^*_2 g^2))$ is given by $g^0=   J_{\Lambda_1}^T J_{\Lambda_2}^T g^2 J_{\Lambda_1}J_{\Lambda_2}$. Note that the matrix $g^0_{p}$ depends on the chosen point $p \in \mathbb{R}^3$ through the Jacobians $J_{\Lambda_1}$ and $J_{\Lambda_2}$. By \Cref{rem:rank_no_bottlenecks}, the matrix $g^0$ is of rank $2$, since $\Ker(g^0_p)$ is a one dimensional subspace of $T_{p} \mathbb{R}^3$; In general the particular subspace $\Ker(g^0_{p})$ of $T_{p} M$ depends on $p$. 
		In particular this means that the classes of equivalence are not straight lines but curves. Furthermore, we know from \cite{BeMa21} that there is a smooth vector field $V: \mathbb{R}^3 \rightarrow T \mathbb{R}^3$ such that $\Ker(g^0_{p}) = \Span\{V_{p}\}$. Since the points of a class of equivalence are on the image of a null curve, the class of equivalence of a point $p \in \mathbb{R}^3$ is given by the points of a curve $\gamma:[-\delta_1,\delta_2] \rightarrow \mathbb{R}^3$ solution of the following Cauchy problem
		\begin{equation}
		\begin{cases}
		\dot{\gamma}(s) = V(x(s),y(s),z(s))\\
		\gamma(0) = p
		\end{cases}
		\end{equation}
	\end{example}
	
	\begin{example}\label{ex:example_metric_ODE}
		Consider $M_0 = (0,+\infty) \times (0,+\infty)$ with the metric $g^0$ represented in Cartesian coordinates $(x,y)$ by 
		\begin{equation}
		g^0_x = 
		\begin{pmatrix}
		x^2 & x \\
		x & 1
		\end{pmatrix}
		\end{equation}
		The eigenvalues of $g^0_x$ are $\lambda_1 = 0$ and $\lambda_1 = 1+x^2$, with eigenvectors $v_1 = (1,-x)$ and $v_2 = (1,x)$. Notice that both the eigenvalues and the eigenvectors are smooth functions of the coordinates: in general \cite{Kato66,Rell69} the eigenvalues and the eigenvectors of matrices depending on some parameters are not even continuous functions of the parameters.
		Let $p=(p_x,p_y) \in M_0$, the points of $M_0$ in the class of equivalence $[p]$ - that in in this case is a one dimensional manifold - lie on the image of a curve $\gamma$ passing through $p$ and such that its tangent vector is in $\Ker(g_0)$ at any point. For example, we can consider the curve $\gamma$ satisfying
		\begin{equation}
		\begin{cases}
		\dot{\gamma}_x (s) = 1 \\
		\dot{\gamma}_y (s) = -x\\
		\gamma_x(0) = p_x\\
		\gamma_y(0) = p_y
		\end{cases}
		\end{equation}
		Solving the system we obtain the curve $\gamma(s)=\left(s+p_x,-\dfrac{s^2}{2}+p_y\right)$, $s \in \mathbb{R}$. The image of this curve is the set of the points of $M_0$ satisfying $y=-\dfrac{1}{2}\left( x-p_x \right)^2+p_y$, namely a parabola.
	\end{example}
	
\begin{figure}[h!]
		\centering
		\includegraphics[scale=1]{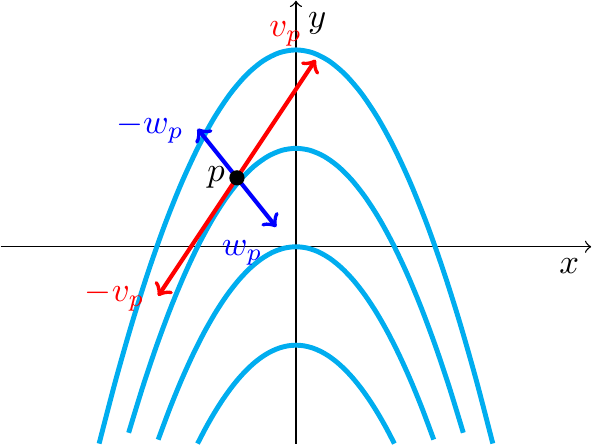}
		\caption{A manifold $M_0$ which is foliated by its classes of equivalence (the curves in cyan). Given a point $p$, by proceeding in the direction of the null vectors $\pm v_p$ we stay on the class of equivalence $[p]$. By proceeding in the direction of a non-null vector $\pm w_p$, we arrive to another class of equivalence.}
		\label{fig:movingClasses}
	\end{figure}

	We know from \cite{BeMa21} that for every point $p \in M_0$ there is a decomposition of $T_p M_0$ as the direct sum $T_p V \oplus T_p H$, with the vector space $T_p V$ spanned by the null eigenvectors of the pullback of $g^{(n)}$ on $M_0$ (which we recall that can be degenerate) and $T_p H$ which is the span of the non-null eigenvectors. Therefore, moving in the direction of a non-null vector, we are changing equivalence class (see \Cref{fig:movingClasses}). In this case, the pseudolength of a curve $\gamma$ from $p \in M_0$ to $q \in M_0$ with $[p] \neq [q]$ corresponds with the length of $\gamma / \sim$ in $M_0 / \sim$.
	
	For simplicity we present a toy example with a simple linear function, but the application to neural network maps is straightforward, as we shall see in \Cref{sec:SiMExp}.
	
	\begin{example}\label{ex:changing_equivalence_class}
		Consider a linear function $f:\mathbb{R}^3\to\mathbb{R}^2$, which in the canonical basis was represented by the matrix
		\begin{equation}
		\begin{pmatrix}
		1 & 2 & 2 \\
		3 & 1 & 5
		\end{pmatrix}
		\end{equation}
		Suppose to endow $\mathbb{R}^2$ with its canonical metric $g$ and to compute the pullback metric $h$, which is given by
		\begin{equation}
		\begin{pmatrix}
		10 & 5 & 17 \\
		5 & 5 & 9 \\
		17 & 9 & 29
		\end{pmatrix}.
		\end{equation}
		Every point $p \in \mathbb{R}^3$  $T_p V = \Ker(h)=\Span\{(8,1,-5) \}$ (see \cite[Example 5]{BeMa21} and that the equivalence class of $p$ is the straight line passing through $p$ and parallel to the vector $(8,1,-5)$. Now we want to  proceed in a direction changing the equivalence class following a vector in $T_p H$. First, we need to build a basis of $T_p H$. After short computation, a possible basis is given by the two following non-null eigenvectors
		\begin{equation}
		v_1 = \left( \frac{52+\sqrt{394}}{55}, -\frac{8\sqrt{394}+141}{55},1 \right)^\top,\, v_2 = \left( - \frac{52-\sqrt{394}}{55}, \frac{8\sqrt{394}-141}{55},1 \right)^\top.
		\end{equation}
		Given a point $p$, proceeding in the direction of any linear combination of $v_1$ and $v_2$ lead us in a point $q$ in another equivalence class. Indeed, if $q = p + \alpha v_1 + \beta v_2$ with $\alpha,\beta \in \mathbb{R}$, $Aq = Ap + \alpha A v_1 + \beta A v_2 = A p + \lambda_1 v_1 + \lambda_2 v_2 \neq Ap$, where $\lambda_{1,2} = 22 \pm \sqrt{394}$ are the eigenvalues of $v_{1,2}$ respectively. By definition $[p] \neq [q]$.
	\end{example}
	The procedure just described is the idea behind the SiMExp algorithm presented in \Cref{sec:SiMExp}. A possible application of this idea is the following. Suppose we have a binary classifier realized through a neural network $\mathcal{N}: M_0 \rightarrow M_1$ with $M_0=(a,b) \times (a,b)$, with $a < b$ two real numbers and $M_1 = (0,1) \subset \mathbb{R}$. This classifier is telling us the probability a point $p \in M_0$ satisfies a certain property. Suppose the $\mathcal{N}(p) = 0.92$. We want to find a region containing $p$ in $M_0$ such that the said property is satisfied between $0.90$ and $0.94$.  
	With a combination of the proposed SiMEC and SiMExp algorithms we can solve this problem. In general, given $\varepsilon>0$, this procedure allow us to numerically build the connected component of $\mathcal{N}^{-1}([\mathcal{N}(p)-\varepsilon,\mathcal{N}(p)+\varepsilon])$ containing $p$ reconstructing the foliation of $M_0$ induced by the equivalence relation. 
	
	\section{The SiMEC and SiMExp algorithms for 1D cases}
	\label{sec:algos}
	In this section we introduce the Singular Metric Equivalence Class (SiMEC)  and the Singular Metric Exploring (SiMExp) algorithms: the former builds one dimensional equivalence classes with the former, whilst the latter is employed to move from an equivalence class to another. We suppose that the manifold $M_n$ in the Neural Network defined as in, \Cref{def:smoothNN} is equipped with a Riemannian metric.
	\subsection{The SiMEC algorithm for 1D equivalence classes}
	As starting point, recall that in the case in which $M=(a_1,b_1)\times\dots \times(a_n,b_n)$, with $-\infty \leq a_i < b_i \leq +\infty$, we can identify $T_pM$ with $\mathbb{R}^n$, in other words we can use the same basis for every tangent space, in this case identified with the canonical basis of $\mathbb{R}^n$, to give the same reference frame over each point $p$.
	In the previous section we stated that following a null curve starting from a point $p \in M_0$ allows one to remain in the class of equivalence of $[p]$, whose points $q$ satisfies $\mathcal{N}(q)=\mathcal{N}(p)$. If $\dim(\Ker(g_0))=1$ then the equivalence classes are curves in $M_0$. In a non-linear regression task, with a training dataset given by the pairs $\{(T_i,O_i)\}$, the loss function is of the form
	\begin{equation}
	\mathcal{L} = \sum_i \| \mathcal{N}(T_i)-O_i \|^2
	\end{equation}
	with $\| \cdot \| $ being the Euclidean norm. This suggests us to choose $g^n$ equal to the standard Euclidean metric. This choice of the metric should be a good choice also for classification problems. Consider a classifier with $m$ classes and let $z_1,\cdots,z_m$ be the coordinates of the output manifold. Since they represent probabilities, they satisfy $\sum_{i =1}^m z_i = 1$. Therefore, $\mathcal{N}(M_0)$ is a ($m-1$)--dimensional submanifold embedded in the Euclidean space $M_n$, whose metric is induced by that of $M_n$. As a consequence, if we want to compute the singular metric $g^0$, we can simply compute the pullback of the canonical Euclidean metric of $M_n$ through the neural network map: this amounts to employ \eqref{eq:metric_pullback} where $g^M$ is $g^0$, $F$ corresponds to  $\mathcal{N}$, $x^i$ and $x^j$ are the coordinates in $M_0$ and finally $g^N$ is the metric on the output manifold. We note that this quantity can be computed via automatic differentiation (see \Cref{rem:autodiff}).
	
	The SiMEC algorithm yields a polygonal approximation of the image of a 1D null curve passing through $p \in M_0$. From the discussion in \Cref{subsec:singular_riemannian_approach_to_nn}, we can build a polygonal approximation of a 1D null curve as follows. Starting from $p$, we diagonalize the pullback $g^0_p$, obtaining the eigenvector $w_0$ associated to the null eigenvalue and the eigenvector $w_+$ associate to the positive eigenvalue. We proceed in a the direction $w_0$ with a step size of length $\delta$ and then repeat this procedure, obtaining a sequence of points $p, p_1, \cdots, p_K$ that are the vertices of the said polygonal. However, even if the core of this idea is correct, there are some geometric and numerical issues to discuss.
	
	From a numerical point of view we have to keep in mind that there are also other numerical errors introduced by the machine, in addition to the fact that we are approximating a curve by means of a piecewise line. The main effect of such numerical errors lies in the computation of the eigenvalues, because of the following related problems:
	\begin{enumerate}[label=\arabic*)]
		\item Even with a non-negative symmetric matrix the machine could yield some negative eigenvalues, although small in absolute value.
		\item The computer considers an eigenvalue as zero if its absolute value is less than the machine epsilon. 
	\end{enumerate}
	The first point has an important consequence: We cannot use the pseudolenght to check if we are approximating well a null curve well, since the argument of the square root may be a negative number. However, we can overcome this issue using the energy of a curve or computing an estimate from above of the pseudolenght as follows
	\begin{equation*}
	L(\gamma) \leq \int_a^b \sqrt{|g_{\gamma(s)}(\dot{\gamma}(s),\dot{\gamma}(s))|} ds
	\end{equation*} 
	
	The second point, along with the other approximations issues, yields that we are not proceeding toward a point lying exactly on $[p]$, but we are only approximating the class of equivalence introducing an error at every step; We can control this error computing the energy of a curve starting from $p$ and ending at the point $p_k$ obtained at the $k-th$ step or using the bound of the pseudolenght given above.
	
\begin{figure}[h!]
		\centering
		\includegraphics[width=.4\textwidth]{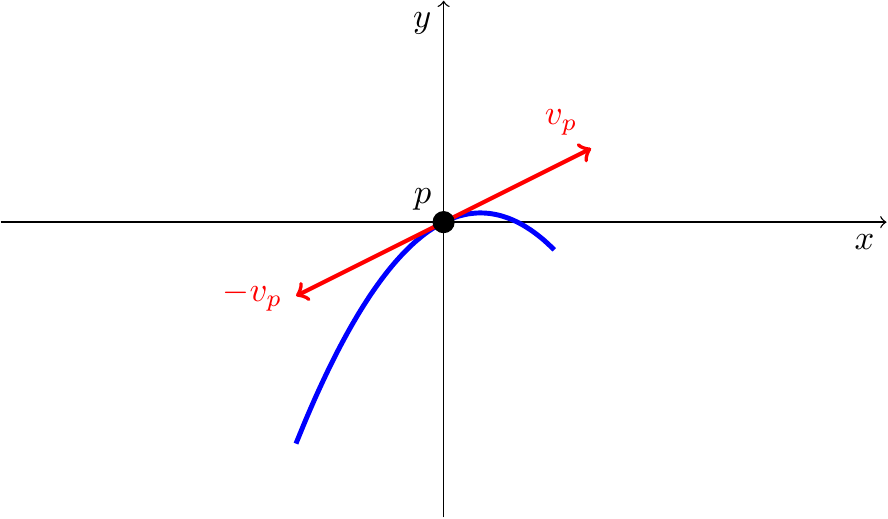}
		\caption{Both $v_p$ and $-v_p$ are normalized eigenvectors at $p$. Sometimes the numerical algorithms computing the eigenvectors of the metric return $v_p$, other times $-v_p$. If we want to consistently proceed forward or backward along a polygonal approximating the curve of an equivalence class built as described above, we need to pick $v_p$ or $-v_p$ accordingly.}\label{fig:eigDirs}
	\end{figure}

	From a geometric point of view, we must discuss a problem related to the choice of an eigenvector of $g^0$ over a point $p$. In general the eigenvectors of a symmetric matrix depending on some parameters are not continuous functions of the parameters \cite{Kato66,Rell69}. However, by \cite{BeMa21}, there is a smooth vector field $X$ which is a basis of $\Ker(g^0_p)$ for every $p \in M_0$, namely $\Span(X_p)=\Ker(g^0_p) \ \forall p \in M_0$. In particular, we can factorize the vector field as $X(p)=f(p)v_p$, with $v_p$ a null eigenvector of $g_0(p)$ and $f$ a smooth function over $M_0$. Since we repeat the procedure above for a finite number of points $\{p_k\}$, we can choose $f$ to be any polynomial of degree high enough such that $f(p_k)=\pm 1$ for every $p_k$ in the sequence of points defining the polygonal (see \Cref{fig:eigDirs}).
	\begin{remark}
		It is necessary to assume that $f(p_k)=\pm 1$ and not only $f(p_k)=1$ also for another reason, related to the numerical algorithms computing the eigenvectors. Suppose that the level sets of a neural network map $\mathcal{N}$ are circles, see for example the first numerical test in \ref{sec:numexp}. Starting from a point $p_0$, we begin to approximate an arc of circle, with the eigenvectors of each step pointing always in the counter--clockwise direction. It may happen that arrived at a certain point $p_l$, the numerical algorithm employed to find the eigenvectors yields the unit vector pointing clockwise instead of that pointing in the opposite direction, as in the previous steps. This means that we begin to go back toward the starting point $p$, eventually passing through the point proceeding clockwise. On the other side of $p$, the same phenomenon may happen again, with the result that we are only able to approximate a part of the level curve. A solution to this problem is to make sure to proceed always in the same direction, fixing $f(p_k)=\pm 1$ accordingly step by step. For example we can do this computing the angle $\theta$ between the eigenvectors $v_k$ and $v_{k+1}$ of two consecutive steps and then taking $+v_{k+1}$ if $-\pi/2 \leq \theta \leq \pi/2$ or $-v_{k+1}$ in the case $\pi/2 < \theta < 3/2 \pi$.
	\end{remark}
	
	Given the data above, the steps of the SiMEC-1D algorithm to build an approximation of $[p] \subset M_0$ are depicted in \Cref{al:SIMEC1D}.
	
	\begin{algorithm}[H]
		\begin{algorithmic}
			\Require {Choose $p_0 \in M_0$, the direction $v_0$, $\delta>0$, maximum number of iteration $K$.}
			\Ensure  {A sequence of points $\{p_s\}_{s=1,\dots,K}$ approximatively in $[p_0]$; The energy $E$ of the approximating polygonal}
			\State{Initialise the energy: $E\gets 0$}
			\For {$k=1,\dots,K-1$}
			\State {Compute $g_{\mathcal{N}(p_k)}^n$}
			\State {Compute the pullback metric $g^0_{p_k}$ trough \Cref{eq:metric_pullback}}
			\State {Diagonalize $g^0_{p_k}$ and find the eigenvectors $w_0,w_+$}
			\State {$v_{k} \gets w_0$}
			\If {$v_{k} \cdot v_{k-1}<0$}
			\State {$v_{k} \gets-v_{k}$}
			\EndIf
			\State {Compute the new point $p_{k+1} \leftarrow p_{k}+ \delta v_{k}$}
			\State {Add the contribute of the new segment to the energy $E$ of the polygonal}
			\EndFor
		\end{algorithmic}
		\caption{The SiMEC 1D algorithm.}\label{al:SIMEC1D}
	\end{algorithm}
	
\begin{figure}[h]\centering
		\begin{subfigure}[t]{0.45\linewidth}
			\centering
			\includegraphics[width=0.5\textwidth]{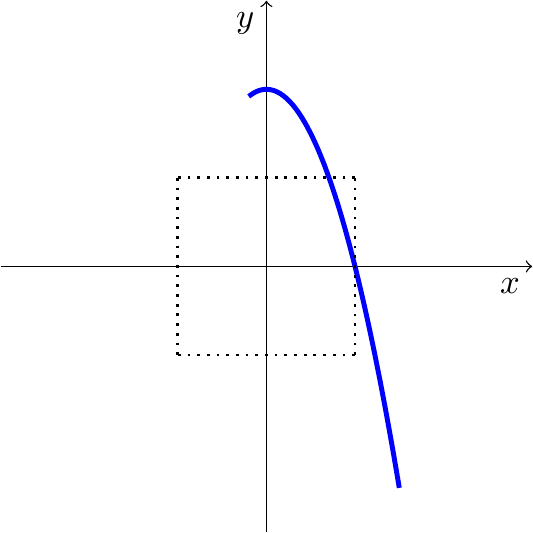}
			\caption{The actual class of equivalence. $(0,1) \times (0,1)$ is the region containing the training data.} 
			\label{fig:example_boundary_1} 
		\end{subfigure}\hfill
		\begin{subfigure}[t]{0.45\linewidth}
			\centering
			\includegraphics[width=0.5\textwidth]{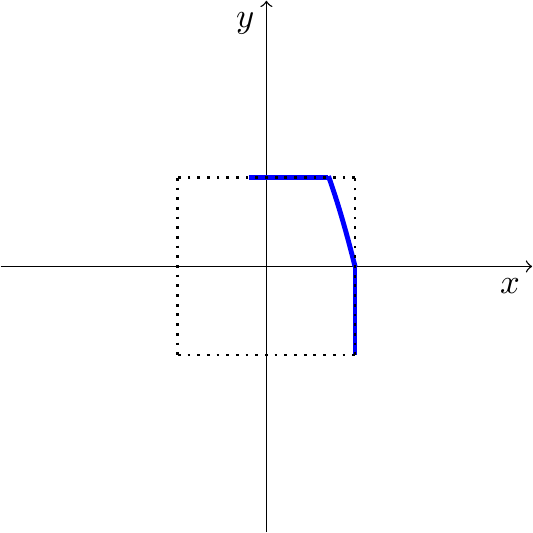}
			\caption{The approximation of the class of equivalence obtained projecting on the square $(0,1) \times (0,1)$.} 
			\label{fig:example_boundary_2} 
		\end{subfigure} 
		\caption{Suppose the equivalence class of the neural network is the arc of parabola represented in \Cref{fig:example_boundary_1}, but the points of the dataset employed to train the neural network lie in the dotted square $(0,1) \times (0,1)$. Approximating the equivalence class projecting on the square yields the blue curve in \Cref{fig:example_boundary_2}, which is a poor approximation of the original parabola.}
	\end{figure}

	Note that on the machine we are always working on sets of the form $(a,b)^n$, for a suitable $n \in \mathbb{N}$. The region $\mathcal{H}$ in which the points of the dataset lie is usually a proper subset of $(a,b)^n$. For example, in the case the training data are normalized, $\mathcal{H}$ is an hypercube of the form $(0,1)^n$ or $(-1,1)^n$. In some cases data have sense only inside this hypercube, both because of some constraint on their values (e.g. a volume cannot be negative) and because outside $\mathcal{H}$ the neural network map does not approximate any meaningful function. 
	\begin{remark}
		\label{rem:autodiff}
		The computation of $g_{\mathcal{N}(p_k)}^n$ in \Cref{al:SIMEC1D} requires the iterative application of \eqref{eq:metric_pullback} (see \cite[Section 2.2.]{BeMa21}). From a numerical point of view, its computation employs the automatic differentiation procedure, using the built--in functions for the derivation wrt to the weights and for the derivation of the activation functions.
	\end{remark}
	However, running the SiMEC algorithm may lead to reach points outside $\mathcal{H}$. To be sure that the points produced by the algorithm make sense, we need either to stop when we hit the boundary of $\mathcal{H}$ or to project the resulting points to a meaningful region (see \Cref{fig:example_boundary_1}). A concrete example is the MNIST dataset - whose data are $28 \times 28$ images: A point can only be in $[0,255]^{28 \times 28}$. A pixel cannot assume, for example, a value of $257$, therefore we must be sure to remain in the interval $[0,255]$ for every point of the $28 \times 28$ image. A possible solution to this problem is to project over the $784$-dimensional hypercube $[0,255]^{784}$. 
	
	To take into account these problems, we propose the modification to SiMEC--1D in \Cref{al:SIMEC1Dmod}. Whether to project on $\mathcal{H}$ or stop when hitting the boundary $\partial \mathcal{H}$ is a choice to be made case by case.
	
	\begin{algorithm}[H]
		\begin{algorithmic}
			\Require {Choose $p_0 \in M_0$, the direction $v_0$, $\delta>0$, maximum number of iteration $K$.}
			\Ensure  {A sequence of points $\{p_s\}_{s=1,\dots,K}$ approximatively in $[p_0]$; The energy $E$ of the approximating polygonal}
			\State{Initialise the energy: $E\gets 0$}
			\For {$k=1,\dots,K-1$}
			\State {Compute $g_{\mathcal{N}(p_k)}^n$}
			\State {Compute the pullback metric $g^0_{p_k}$ trough \Cref{eq:metric_pullback}}
			\State {Diagonalize $g^0_{p_k}$ and find the eigenvectors $w_0,w_+$}
			\State {$v_{k} \gets w_0$}
			\If {$v_{k} \cdot v_{k-1}<0$}
			\State {$v_{k} \gets-v_{k}$}
			\EndIf
			\State {Compute the new point $p_{k+1} \leftarrow p_{k}+ \delta v_{k}$}
			\State {Project on the hypercube $\mathcal{H}$ or stop when hitting the boundary $\partial \mathcal{H}$}
			\State {Add the contribute of the new segment to the energy $E$ of the polygonal}
			\EndFor
		\end{algorithmic}
		\caption{Modification of the SiMEC 1D algorithm taking in account the problem of exiting the meaningful region.}\label{al:SIMEC1Dmod}
	\end{algorithm}

	If we do not halt the algorithm when we hit $\partial \mathcal{H}$, the projection on the hypercube is a source of errors in the approximation of a null curve, since we are not projecting on null directions. Therefore we expect to see an additional increase of the energy of the curve every time we perform the projection.
	
	\subsection{Changing equivalence classes : The SiMExp algorithm.}
	\label{sec:SiMExp}
	In this section we discuss the Singular Metric Exploring algorithm, allowing one to pass from a given equivalence class to a near one. Under the hypotheses of \Cref{sec:geometric_preliminaries} and \Cref{sec:geometric_approach_to_nn}, we know from \cite{BeMa21} that for every point $p \in M_0$ there is a decomposition of $T_p M_0$ as the direct sum $T_p V \oplus T_p H$, with the vector space $T_p V$ spanned by the null eigenvectors and $T_p H$ which is the span of the non-null eigenvectors. Therefore, moving in the direction of a non-null vector, we are changing equivalence class. We limit ourselves to describe the SiMExp algorithm for function from $\mathbb{R}^2$ to $\mathbb{R}$; From a geometric point of view the generalization to the case $\mathbb{R}^n \rightarrow \mathbb{R}^{n-1}$ is straightforward, as it is enough to consider any of the non-null eigenvectors to change the class of equivalence. However, if the dimension input manifold $M_0$ is not low enough, proceeding in the same way is very demanding form a computational point of view. Let $p \in M_0$ with $\mathcal{N}(p) = \alpha$. The SiMExp algorithm allows us to find a set of points $S\subset M_0$, containing $p$, such that $\mathcal{N}(S) \subseteq [\alpha - \varepsilon, \alpha + \varepsilon]$. As for the SiMEC algorithm, there is the problem of exiting the region $\mathcal{H}$ in which the training data lie.
	\begin{algorithm}
		\begin{algorithmic}
			\Require {Choose $p_0 \in M_0$, direction $v_0$, $\delta>0$, tolerance parameter $\varepsilon$.}
			\Ensure  {A sequence $\{p_s\}_{s\in\mathbb{N}}$ of points such that $\mathcal{N}(p_s) \subseteq [\mathcal{N}(p_0)-\varepsilon,\mathcal{N}(p_0)+\varepsilon]$; The length $\ell$ of the approximating polygonal}
			\State{Initialise the length: $\ell\gets 0$}
			\State{$k\gets 1$}
			\Repeat 
			\State {Compute $g^n_{\mathcal{N}(p_k)}$}
			\State {Compute the pullback metric $g^0_{p_k}$ trough \Cref{eq:metric_pullback}}
			\State {Diagonalize $g^0_{p_k}$ and find the eigenvectors $w_+,w_0$}
			\State {$v_{k} \gets w_+$}
			\If {$v_{k} \cdot v_{k-1}<0$}
			\State {$v_{k} \gets-v_{k}$}
			\EndIf
			\State {Compute the new point $p_{k+1} \leftarrow p_{k}+ \delta v_{k}$}
			\State {$k\gets k+1$}
			\State {Project on the hypercube $\mathcal{H}$ or stop when hitting the boundary $\partial \mathcal{H}$}
			\State {Add the contribute of the new segment to the length of the polygonal}
			\Until{$|\mathcal{N}(p_k)-\mathcal{N}(p)| \leq \varepsilon$}
		\end{algorithmic}
		\caption{SiMExp-1D algorithm taking in account the problem of exiting the meaningful region.}\label{al:SIMEXP1}
	\end{algorithm}
	
	To find an approximation of the set $S$ of all the points in $M_0$, with $p \in S$, such that $|\mathcal{N}(p_k)-\mathcal{N}(p)| \leq \varepsilon$, we can use a combination of 
	SiMExp and SiMEC. In order to be sure to stop after a finite number of steps, we run both SiMExp and SiMEC until we reach the boundary of the region $\mathcal{H}$. In particular this is tantamount to build the connected component of $\mathcal{N}^{-1}([\mathcal{N}(p)-\varepsilon,\mathcal{N}(p)+\varepsilon])$.
	\begin{algorithm}[H]
		\begin{algorithmic}
			\Require {Choose $p_0 \in M_0$, $\delta>0$, tolerance parameters $\varepsilon$, $\tilde{\varepsilon}\ll 1$.}
			\Ensure  {A set $\mathcal{S} = \{p_\ell\}_{\ell\in\mathbb{N}}$ of points such that $\mathcal{N}(p_\ell) \subseteq [\mathcal{N}(p_0)-\varepsilon,\mathcal{N}(p_0)+\varepsilon]$}
			\State{Initialise the length: $\ell\gets 0$}
			\State {Compute $g^n_{\mathcal{N}(p_0)}$}
			\State {Compute the pullback metric $g^0_{p_0}$ trough \Cref{eq:metric_pullback}}
			\State {Diagonalize $g^0_{p_0}$ and find the eigenvectors $w_+,w_{0}$}
			\Repeat 
			\State {$\{p_s\}\gets$ SiMEC($p_{k},\delta,K$)}
			\State {$\mathcal{S}\hookleftarrow \{p_s\}$}
			\State {$p_{k+1}\gets$ SiMExp($p_{k},w_+,\delta,\tilde{\varepsilon}$)}
			\State {$k\gets k+1$}
			\Until{$|\mathcal{N}(p_k)-\mathcal{N}(p)| \leq \varepsilon$}
			\State {$k\gets 0$}
			\Repeat 
			\State {$\{p_s\}\gets$ SiMEC($p_{k},\delta,K$)}
			\State {$\mathcal{S}\hookleftarrow \{p_s\}$}
			\State {$p_{k+1}\gets$ SiMExp($p_{k},-w_+,\delta,\tilde{\varepsilon}$)}
			\State {$k\gets k+1$}
			\Until{$|\mathcal{N}(p_k)-\mathcal{N}(p)| \leq \varepsilon$}
		\end{algorithmic}
		\caption{Algorithm approximating the set $S$ of all the points in $M_0$, $p \in S$, such that $|\mathcal{N}(p_k)-\mathcal{N}(p)| \leq \varepsilon$}\label{al:SIMEXP2}
	\end{algorithm}
	
	\section{Numerical experiments}
	\label{sec:numexp}
	This section is devoted to asses the performance of the proposed algorithm and the quality of the provided results. The numerical experiments were carried on a machine running Ubuntu 20.04, equipped with an eight-cores i7-10700K processor providing 16 logic CPUs, 24 GB of RAM memory and a GeForce RTX 2060 graphic card. The neural networks were trained using Keras with Tensorflow 2.3.1 as backend. We implemented the SiMEC and SiMExp algorithms in a neural network built from scratch in C++. The code is available at \url{http://github.com/alessiomarta/simec-1d-test-code}.
	
	\subsection{Numerical experiments for SiMEC-1D}
	\label{sec:numerical_experiments}
	In this section we present some numerical experiments in which we apply the SiMEC-1D algorithm to non-linear regression problems. In all these experiments the neural network is learning a function from $\mathbb{R}^2$ to $\mathbb{R}$, restricted to a suitable subset: The region in which we generate the features employed for the training. The training of the neural network has been done using 
	
	\paragraph{Learning compact equivalence classes} In this numerical experiment, we generated a cloud of $2000$ points lying on the surface $z=e^{x^2+y^2-2}$, with $(x,y)$ randomly generated $(-1,1)\times (-1,1)$ using the uniform distribution. Then we trained the neural network 
	\begin{equation}
	\begin{tikzcd}
	\mathbb{R}^{2} \arrow[r, "\Lambda_1"] & \mathbb{R}^{5} \arrow[r, "\Lambda_2"] & \mathbb{R}^{5} \arrow[r,"\Lambda_3"]  & \mathbb{R}
	\end{tikzcd}
	\end{equation}
	to learn the function $(x,y) \mapsto e^{x^2+y^2-2}$ using the cloud of points above. The maps $\Lambda_i$ employ sigmoid functions as activations. Since this is a (non-linear) regression problem, we trained the neural network choosing as loss function the mean squared error. To perform the training we employed the Adam optimization algorithm with a batch of $512$ element. We run the training process for $5000$ epochs reaching a mean square error of $1.0567 \cdot {10}^{-5}$ for the training data. We considered the first half of the points as the training dataset and the second half as the validation dataset. 
	
	\begin{figure}[h]
		\centering
		\begin{subfigure}[t]{.475\textwidth}
			\includegraphics[width=1.\textwidth]{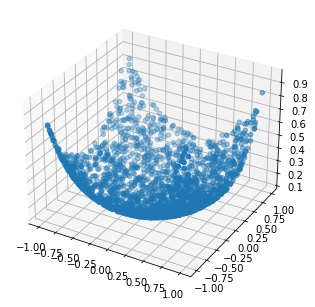}
			\caption{Point cloud.}
			\label{fig:test1}
		\end{subfigure}%
		\begin{subfigure}[t]{.475\textwidth}
			\includegraphics[width=1.\textwidth]{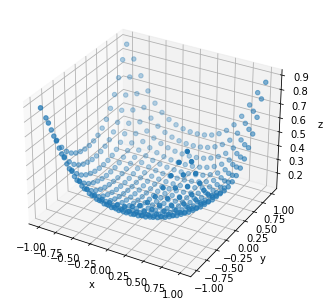}
			\caption{Learned function.}
			\label{fig:exp_1_learned_surface}
		\end{subfigure}
		\caption{Left panel: cloud point for the training of the neural network, half of the points are used for the training and half for the validation. Right panel: plot of the learned function over equispaced points in the interval $[-1,1]\times[-1,1]$.}
	\end{figure}
	
\begin{figure}[h]
		\centering
		\newcommand{\factor}{0.19}
		\begin{subfigure}[t]{\factor\textwidth}
			\centering
			\includegraphics[width=1.\linewidth]{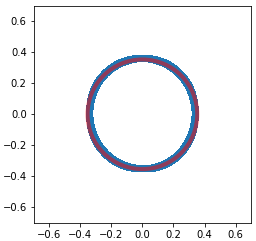}
			\caption{$\delta=2 \cdot 10^{-6}$.}
			\label{fig:test1ok}
		\end{subfigure}%
		\begin{subfigure}[t]{\factor\textwidth}
			\centering
			\includegraphics[width=1.\linewidth]{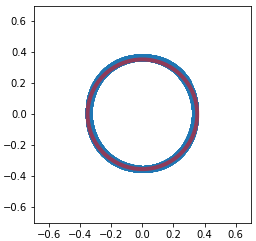}
			\caption{$\delta = 1 \cdot 10 ^{-3}$.}
			\label{fig3:c}
		\end{subfigure}
		\begin{subfigure}[t]{\factor\textwidth}
			\centering
			\includegraphics[width=1.\linewidth]{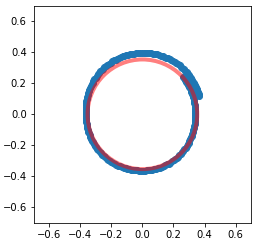}
			\caption{$\delta = 1.25 \cdot 10 ^{-2}$}
			\label{fig3:b}
		\end{subfigure}
		\begin{subfigure}[t]{\factor\textwidth}
			\centering
			\includegraphics[width=1.\linewidth]{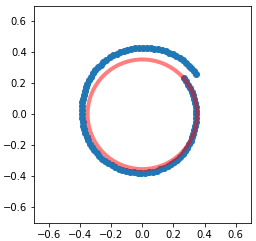}
			\caption{$\delta = 2.5 \cdot 10 ^{-2}$.}
			\label{fig3:a}
		\end{subfigure}
		\begin{subfigure}[t]{\factor\textwidth}
			\centering
			\includegraphics[width=1.\linewidth]{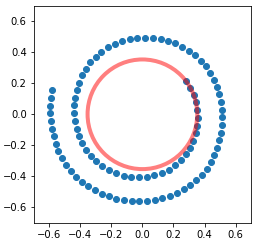}
			\caption{$\delta=5 \cdot 10^{-2}$.}
			\label{fig:poor_approx_spiral}
		\end{subfigure}
		\caption{Plots of the class of equivalence for different values of $\delta$ (in blue) versus the real level set, the red circle centered at the origin and passing through $P$. The higher the value for $\delta$, the poorer the reconstruction.}
	\end{figure}

	We apply the SiMEC-1D algorithm with $\delta=2 \cdot 10^{-6}$ to build the equivalence class of $P=(.25,.25)$. The level sets of $f$ are circles, therefore if the network properly learned the function, we should obtain equivalence classes that are approximatively circles. Computing the output of the neural network on these points yields $0.154072$ for the whole polygonal approximating the equivalence class. Therefore the approximation errors are less than order $10^{-6}$. The final energy, after $1.5 \cdot 10^6$ iterations is $1.01552 \cdot 10^{-21}$. Since the whole class of equivalence of this particular point is in the region $(0,1) \times (0,1)$ there is no need to project on the square or to halt the iterations once we hit the boundary. The resulting equivalence class is shown in \Cref{fig:test1ok}.

	We try to build the equivalence class of the point $P$ with $\delta = 5 \cdot 10^{-2}$: since $\delta$ is larger than before, we need fewer iterations to build the whole curve. $150$ iterations yields the approximation depicted in \Cref{fig:poor_approx_spiral}. The points in blue are the vertices of the approximating polygonal. This time the polygonal is a good approximation only for the first few iterations, and then it starts to degenerate. This behaviour is due to the fact that $\delta$ is not low enough and the error committed computing $p_{k+1} = p_k + \delta v_k$ is not negligible as we move away enough from the true level curve to compute -- in the next iteration -- the eigenvectors along another level curve which is far enough from the original one to give perceptible differences. As we lower the value of $\delta$, the approximations get better.
	
\begin{figure}[h]
		\centering
		\begin{subfigure}[t]{0.5\textwidth}\centering
			\includegraphics[width=.75\textwidth]{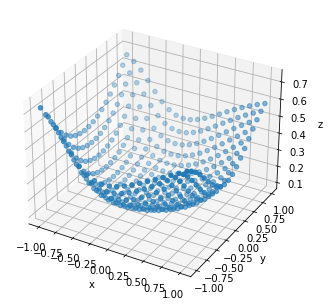}
			\caption{Learned function.}
			\label{fig:num_exp_1_poor_surface}
		\end{subfigure}\hfill
		\begin{subfigure}[t]{0.5\textwidth}\centering
			\includegraphics[width=1.\textwidth]{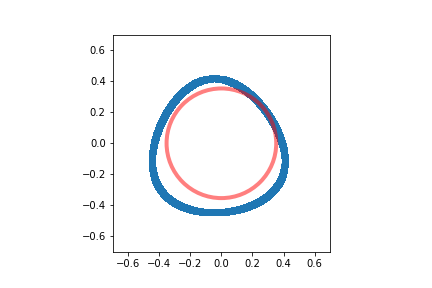}
			\caption{Equivalence class (blue) VS the real level curve (red).}
			\label{fig:num_exp_1_poor_eq}
		\end{subfigure}
		\caption{Left panel: the degree of approximation achieved by the neural network is not reliable, as it is evident by the 3D plot. Right panel: if the approximation is poor, as one expects the estimation of a equivalence class is poor too. Indeed, the SiMEC algorithm reconstructs the equivalence class of the function learned by the NN.}
	\end{figure}

	The result of the algorithm also depends on the degree of approximation, \emph{i.e.} on the goodness of the fit provided by the NN. Poor approximation of the function to learn leads to poor approximation of the equivalence classes. For example, we trained the previous neural network until it reached a loss $L=0.002$. After $1.5 \cdot 10^6$ iterations, the SiMEC-1D algorithm with $\delta = 2 \cdot 10^{-6}$ produced and egg-shaped curve passing trough the starting point $(0.25,0.25)$ which is a poor approximation of the true level curve. Indeed the learned surface is a poor approximation of the paraboloid given by $z=f(x,y)$ - Compare \Cref{fig:num_exp_1_poor_surface} with \Cref{fig:exp_1_learned_surface}, notice in particular the asymmetry with respect to the $x$-coordinate. 
	The energy of the curve approximating the equivalence class is $2.67 \cdot 10^{-21}$ and the output of the neural network is $0.141256$ over the whole curve: The numerical errors are less than $10^{-6}$; Lower values of $\delta$ leads to the very same curve, since the problem lies in the neural network approximating the original paraboloid poorly.

	\paragraph{Learning non compact equivalence classes} We repeat the same experiment with a network learning the function $e^{x^2+y-2}$, whose level sets are parabolas, which are not compact sets. We built the cloud of points generating random values of $x,y$ in $(0,1) \times (0,1)$. This time, using a neural network which is trained well enough, we do not expect compact equivalence classes, therefore we have to tackle the problem of exiting the region in which the training data lies.
	\begin{figure}[h]
		\centering
		\includegraphics[width=.5\textwidth]{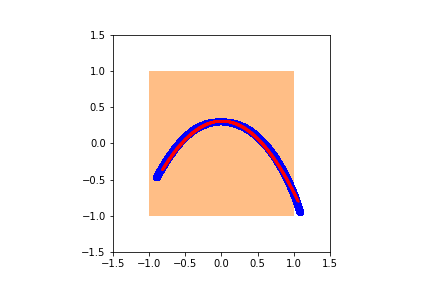}
		\caption{Plot of the class of equivalence (in blue) in a neighbourhood of the point $(0.25,0.25)$. The orange square represent the region in which we generated the training data. The red curve is the real level curve. We generated this plot running the SiMEC-1D algorithm for $15000$ steps in both directions with $\delta=10^{-5}$. For both the direction the final energy is of order $10^{-22}$.} 
		\label{fig:nume_exp_2}
	\end{figure}
	We trained a neural network
	\begin{equation}
	\begin{tikzcd}
	\mathbb{R}^{2} \arrow[r, "\Lambda_1"] & \mathbb{R}^{5} \arrow[r, "\Lambda_2"] & \mathbb{R}^{5} \arrow[r,"\Lambda_3"]  & \mathbb{R}
	\end{tikzcd}
	\end{equation}
	to learn the function $(x,y) \mapsto e^{x^2+y-2}$ using the cloud of points we generated. We chose the mean squared error as loss function, reaching a loss of $3.6337\cdot 10^{-5}$ after $20000$ epochs. To perform the training we employed the Adam optimization algorithm with a batch of $512$ elements. Building the class of equivalence of $(0.25,0.25)$ yields the plot in \Cref{fig:nume_exp_2}.
	
\begin{figure}[htbp]
		\centering
		\begin{subfigure}[t]{0.2\textwidth}
			\centering
			\includegraphics[width=.5\textwidth]{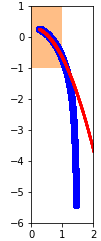}
			\caption{Without projection.}
			\label{fig:longwithout}
		\end{subfigure}\hfill
		\begin{subfigure}[t]{0.2\textwidth}
			\centering
			\includegraphics[width=.5\textwidth]{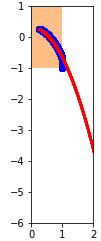}
			\caption{With projection.}
			\label{fig:longwith}
		\end{subfigure}
		\begin{subfigure}[t]{0.25\textwidth}\centering
			\includegraphics[width=1\textwidth]{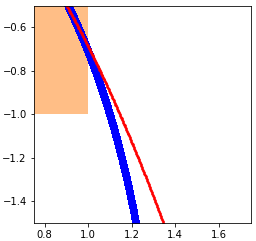}
			\caption{Without projection.}
			\label{fig:shortwithout}
		\end{subfigure}\hfill
		\begin{subfigure}[t]{0.25\textwidth}
			\centering
			\includegraphics[width=1\textwidth]{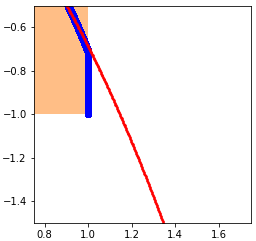}
			\caption{With projection.}
			\label{fig:shortwith}
		\end{subfigure}
		\caption{From left to right: approximation of the equivalence class of $(0.25,0.25)$ without and with projection in the original domain, in the subregions $(0,2)\times(1,-6)$ (first two panels) and in $(0.75,1.65)\times(-1.5,0.5)$. The blue line refers to the polygonal obtained starting from $(0.25,0.25)$ after $600000$ steps. The red line is the true level curve, whilst the orange area is the original domain.}
		\label{fig:num_exp_2_no_projection}
	\end{figure}

	Running the algorithm further, for example from right to left, produce a curve continuing outside the region $\mathcal{H}=(-1,1) \times (-1,1)$. The starting point is still $(0.25,0.25)$. The result is shown in \Cref{fig:num_exp_2_no_projection}.
	As expected, out of the square $(0,1) \times (0,1)$ the quality of the approximation rapidly deteriorates. A solution is to halt the algorithm when the polygonal hits the boundary, obtaining a result as in \Cref{fig:nume_exp_2}. Otherwise, we can project on the square.
	In this case, from a graphical point of view, the polygonal obtained without the projection, see \Cref{fig:longwithout,fig:shortwithout}, is a better approximation of the level curve in a small neighbourhood of the square $(0,1) \times (0,1)$ compared to the one built projecting on the square, see \Cref{fig:longwith,fig:shortwith}. However, in some cases, the variables $x,y$ may lose sense outside $(-1,1) \times (-1,1)$ and we cannot continue outside the square. For example, $y \in (-1,1)$ may refer to a normalized temperature of a classical system originally in $(0 \ K , 273.15 \ K) $: Any value of $y$ below $-1$ correspond to a temperature under the absolute zero, which has no sense. 
	
	We also note that the energy and the pseudolength of a polygonal obtained projecting on the square or exiting the square lose their sense as objects measuring the accuracy of the approximation.
	
\begin{figure}[h!]
		\centering
		\begin{subfigure}[t]{0.45\textwidth}
			\centering
			\includegraphics[width=1\textwidth]{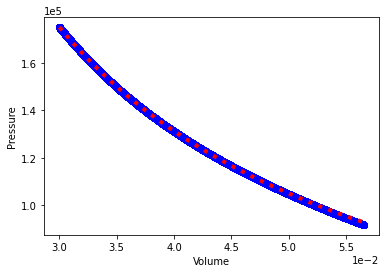}
			\caption{Result achieved using a reliable approximation.}
			\label{fig:thermo1}
		\end{subfigure} \hfill
		\begin{subfigure}[t]{0.45\textwidth}
			\centering
			\includegraphics[width=1\textwidth]{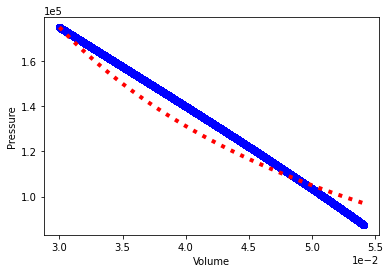}
			\caption{Result achieved using a poor approximation.}
			\label{fig:thermo2}
		\end{subfigure} 
		\caption{Plot of the isothermal curve generated by the algorithm (in blue) and of the real isothermal curve at $631.77 \ K$ (in red). The left panel shows the result obtained by employing a reliable approximation, \emph{i.e.} when the NN is properly trained, while the right panel shows again how a poor approximation leads to have poor results.} 
		\label{fig:thermoTOT}
	\end{figure}

	\paragraph{A thermodynamics problem} The numerical experiment presented in this section relies on a classical physics problem: the estimation of a isothermal curve of a gas. The volume $V$ occupied by $n$ moles of an ideal gas, its pressure $P$ and its temperature $T$ are related through the equation
	\begin{equation}\label{eq:pv=nrt}
	PV = nRT
	\end{equation}
	with being $R \approx 8.314462 J \ mol^{-1} / K^{-1}$ the universal gas constant. Suppose we want to find the temperature of $1$ mole of gas measuring $P$ and $V$. Solving \Cref{eq:pv=nrt} with respect to $T$ yields
	\begin{equation}
	T = \frac{PV}{nR}\label{eq:temp_ideal_gas}
	\end{equation}
	In our numerical experiment we created a dataset -- for fixed number of moles $n=1$ -- randomly generating some triples $(P,V,T)$  with $V$ in the interval $(2.5\cdot 10^{-2} \ m^3 \ ; \ 7.5 \cdot 10^{-2} \ m^3)$ and $P \in (1\cdot 10^{5} \ Pa \ ; \ 2 \cdot 10^{5} \ Pa)$ ; The corresponding values of $T$ are compute using \Cref{eq:temp_ideal_gas}. Then we trained a neural network to predict $T$ from the knowledge of $V,P$. On the base of these premises, we expect the $PV$ plot of a class of equivalence to represent an isothermal curve of the gas. Our neural network is the following, where the maps $\Lambda_i$ are sigmoid activation.
	\begin{equation}
	\begin{tikzcd}
	\mathbb{R}^{2} \arrow[r, "\Lambda_1"] & \mathbb{R}^{5} \arrow[r, "\Lambda_2"] & \mathbb{R}^{10} \arrow[r,"\Lambda_3"]  & \mathbb{R}^{10} \arrow[r,"\Lambda_4"] & \mathbb{R}^{5} \arrow[r,"\Lambda_5"] & \mathbb{R}
	\end{tikzcd}
	\end{equation}
	All the layers use a sigmoid activation function. We trained -- on normalized $P,V,T$ -- the network for $10000$ epochs, reaching a loss $L= 1.9983 \cdot 10^{-06}$. Since we are dealing with a non-linear regression problem, we chose as loss function the mean squared error. The plots in \Cref{fig:thermoTOT} show the curve obtained through the algorithm and the one obtained with \Cref{eq:temp_ideal_gas}. We started from the point $A$ with $V=3 \cdot 10^{-2} \ m^3$, $P = 1.75 \cdot 10^{5} \ Pa$ and we made the algorithm run forward. We expect to see the isothermal curve passing through $A$, namely the isothermal curve of ideal gas at $631.77 \ K$.
	
	The plot of the class of equivalence in \Cref{fig:thermo1} is indeed an approximation of the theoretical isothermal curve. As already observed, the results heavily depends on the value of the loss, \emph{i.e.} on the goodness of the fit. A network approximating \Cref{eq:temp_ideal_gas} poorly yields equivalence classes which are not isothermal curves in the $PV$ plane. For example, a neural network with the same structure as above trained with loss $L=1.1 \cdot 10^{-3}$ yields the equivalence class depicted in \Cref{fig:thermo2} for the point $A$ previously considered.
	
	\subsubsection{Numerical experiments for SiMExp-1D and for coupling SiMEC and SiMExp}
	\label{subsub:numerical_exp_4}
	This section, together with the forthcoming ones, is devoted to asses the performance and the results of SiMExp algorithm and of the combination of both SiMExp and SiMEC.
	
\begin{figure}[h]
		\centering
		\begin{subfigure}{.35\textwidth}
			\centering
			\includegraphics[width=\textwidth]{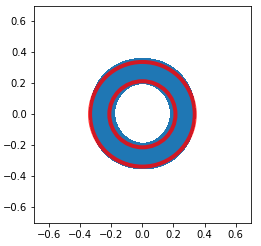}
			\caption{$\delta=10^{-4}$.}
			\label{fig:ann1}
		\end{subfigure}%
		\hspace{0.15\textwidth}
		\begin{subfigure}{.35\textwidth}
			\centering
			\includegraphics[width=\textwidth]{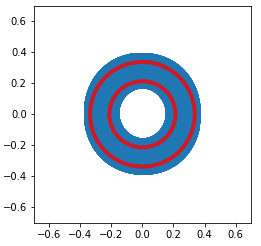}
			\caption{$\delta=10^{-3}$.}
			\label{fig:ann2}
		\end{subfigure}
		\caption{Approximations of the region $S=\mathcal{N}^{-1}([0.147957-\varepsilon, 0.147957+\varepsilon])$ for $\varepsilon=0.05$ and for different values of $\delta$. The red circles are the boundaries of the real annulus obtained from the function $exp(x^2+y^2-2)$. The parameter $\delta$ plays a crucial role in the reconstruction of the region of interest.}
		\label{fig:annTOT}
	\end{figure}

	\paragraph{Learning preimages of compact equivalence classes} This experiment considers again the neural network $\mathcal{N}_1$ trained for the approximation of $(x,y)\to e^{x^2+y^2-2}$. The point $(.2,.2)$ is such that $\mathcal{N}_1(0.2,0.2)=0.147957$: we want hence to find the set of points $S$ in $(0,1) \times (0,1)$ such that $\mathcal{S} \subseteq [0.147957-\varepsilon, 0.147957+\varepsilon]$ -- being the equivalence classes compact subsets, we have $\mathcal{S} = \mathcal{N}^{-1}_1([0.147957-\varepsilon, 0.147957+\varepsilon])$. Running the combination of SiMEC and SiMExp presented in \Cref{al:SIMEXP2} with $\varepsilon=0.05$ produces the set of points depicted in \Cref{fig:num_exp_binary_cloud}. In this case, even with a well--trained network, the parameter $\delta$ plays a crucial role: for a value too large (see \Cref{fig:ann2}) the annulus $\mathcal{S}$ is overestimated, whilst for a smaller value ($\delta=10^{-4}$, \Cref{fig:ann1}) the reconstruction of $\mathcal{S}$ is more precise.
	
	\paragraph{Thermodynamics: learning a family of isothermal curves}
	\begin{figure}[htbp]
		\centering
		\includegraphics[width=.5\textwidth]{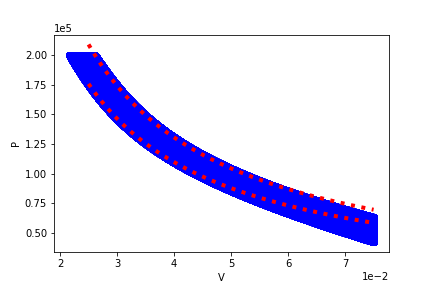}
		\caption{The set in blue is $S=\mathcal{N}^{-1}(0.184-0.03,0.0184+0.03)$. The two dotted curves in red are the the real isothermal curve at $528 \ K$ (below) and $628.9 \ K$ (above).}
		\label{fig:between_isothermal}
	\end{figure} We consider again the thermodynamics application of \Cref{eq:pv=nrt,eq:temp_ideal_gas} and the respective trained neural network $\mathcal{N}_3$. Using \Cref{eq:temp_ideal_gas}, one has that at point $A = (0.03 \ m^3 , 1.75 \cdot 10^{5} \ Pa)$ an isothermal curve at $T=631.77 \ K$. If we use the neural network $\mathcal{N}_3$, we need to normalize also the temperatures. For the cloud of points of the dataset $(2.5\cdot 10^{-2},7.5\cdot 10^{-2})m^3\times (10^5,2\cdot10^5)Pa$ one has that $T_{Max}\approx 1804.2 \ K$, while $T_{min} \approx 300.7 \ K$, by \Cref{eq:temp_ideal_gas}. The normalized value of $T_A$ is $\widetilde{T}_A = 0.184$. If we run \Cref{al:SIMEXP2} with $\varepsilon = 0.03$ -- for $\delta$ small enough -- we should obtain the set $S= \mathcal{N}^{-1}(0.184-0.03,0.0184+0.03)$, corresponding to all the points belonging to the isothermal curves between $528 \ K$ and $628.9 \ K$. The result with $\delta = 10^{-4}$ is shown in \Cref{fig:between_isothermal}. In a neighbourhood of the starting point $A= (0.03 \ m^3 , 1.75 \cdot 10^{5} \ Pa)$ the set produced by the algorithm is a good approximation of the region between the two isothermal curves. Towards the boundary of the region in which we generate the cloud of points the approximation gets worse.

	\paragraph{A classification problem} The last experiment presented in this work regards a classification problem. In this numerical test we generated a dataset in $\mathbb{R}^3$ in the following way: We considered $(x,y) \in (-\pi,\pi)\times(-1,1)$, if a point is such that $y \geq \sin(x)$, then $z=1$, otherwise $z=0$ (see \Cref{fig:sindata1}). As a 2D visual inspection, we depict a point with $z=1$ in red, while points with $z=0$ are coloured in blue (see \Cref{fig:sindata2}).
	\begin{figure}[h!]
		\centering
		\begin{subfigure}[t]{.5\textwidth}
			\centering
			\includegraphics[width=1\textwidth]{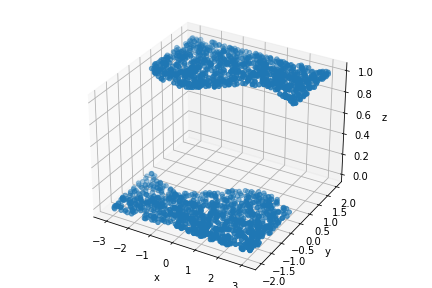}
			\caption{The dataset as a surface.}
			\label{fig:sindata1}
		\end{subfigure}\hfill
		\begin{subfigure}[t]{.5\textwidth}
			\centering
			\includegraphics[width=1\textwidth]{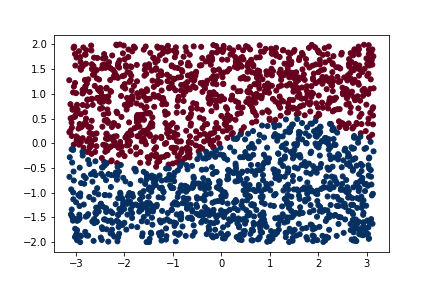}
			\caption{2D coloured visualization of the dataset.}
			\label{fig:sindata2}
		\end{subfigure}
		\begin{subfigure}[t]{.5\textwidth}
			\centering
			\includegraphics[width=1\textwidth]{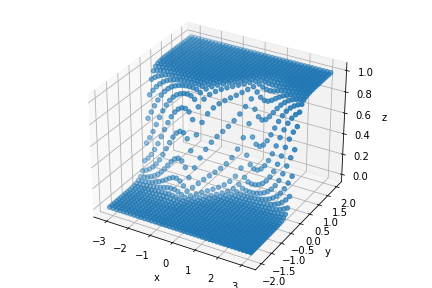}
			\caption{Learned surface by the neural network.}
		\end{subfigure}\hfill
		\begin{subfigure}[t]{.5\textwidth}
			\centering
			\includegraphics[width=1\textwidth]{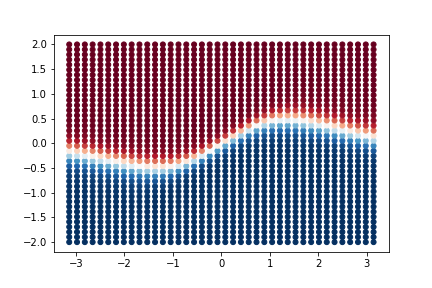}
			\caption{Coloured inspection of the classification done by the NN. Points on the white line are classified as red with a probability of 50\%.}
		\end{subfigure}
		\caption{A classification problem. Top left panel: 3D representation of the dataset, all the points $(x,y)$ such that $y\geq \sin(x)$ have the 3rd coordinate equal to 1. Top right panel: blue dots represent $(x,y)$ points such that $y<\sin(x)$, red dots represent points such that $y\geq \sin(x)$. Bottom left panel: reconstruction of the surface via a neural network: it is clear that the sharp discontinuity is not recovered, since we are employing a smooth neural network. Right bottom panel: coloured representation of the classification, using the red--blue colormap.}
		\label{fig:num_exp_binary_cloud}
	\end{figure}
	
\begin{figure}
		\centering
		\begin{tabular}{cc}
			\begin{tabular}{c}
				\begin{subfigure}{.5\textwidth}
					\centering
					\includegraphics[width=\textwidth]{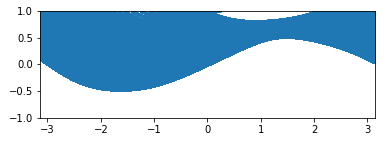}
					\caption{Result with $\delta=1 \cdot 10^{-6}$.}
					\label{fig:bin_foliation_1}
				\end{subfigure}  \\
				\begin{subfigure}{.5\textwidth}
					\centering
					\includegraphics[width=\textwidth]{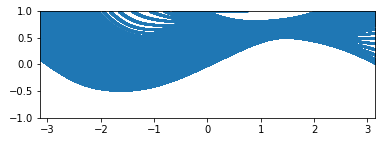}
					\caption{Result with $\delta=2 \cdot 10^{-6}$.}
					\label{fig:bin_foliation_2}
				\end{subfigure}
			\end{tabular} &
			\begin{subfigure}{.45\textwidth}
				\centering
				\includegraphics[width=\textwidth]{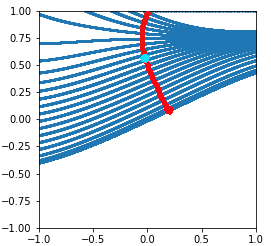}
				\caption{Reconstruction of the foliation of the input manifold near the starting point, in light blue.}
				\label{fig:foliation5e-6}
			\end{subfigure}  
		\end{tabular}
		\caption{Left panels: The set $S=\mathcal{N}^{-1}(0.9,1.0)$ of points which are red with a probability higher than $90 \%$ is built reconstructing the foliation of the input manifold. As $\delta$ increases, the curves of the foliation are more separated; Indeed in \Cref{fig:bin_foliation_2} some of these curves are clearly visible. Right panel: zoom of the reconstruction of the foliation of the input manifold near the starting point, in light blue.}
	\end{figure}

	Training the neural network
	\begin{equation}
	\begin{tikzcd}
	\mathbb{R}^{2} \arrow[r, "\Lambda_1"] & \mathbb{R}^{5} \arrow[r, "\Lambda_2"] & \mathbb{R}^{5} \arrow[r,"\Lambda_3"]  & \mathbb{R}^{5} \arrow[r,"\Lambda_4"] & (0,1)
	\end{tikzcd}
	\end{equation}
	(where $\Lambda_i$ is a sigmoid activation for any $i$) for $20000$ iterations with the Adam algorithm we reached a loss $L = 7.0229 \cdot 10^{-04}$. The output of the neural network is the probability of a point being red. As a first example we want to find the points which are red with a probability higher than $90\%$. To this end, we look for a point whose probability to be red is almost $100 \%$. We chose the point $(0,.5)$, which is red with a probability of $99.9941 \%$. Then we passed $(0,1)$ as input to \Cref{al:SIMEXP2} with $\delta = 10^{-6}$ and $\varepsilon=0.1$. In this way we are considering the interval $(0.9,1)$ as the output of the neural network -- Remember that we cannot overpass $1$, so $\mathcal{N}^{-1}(0.9,1.1)=\mathcal{N}^{-1}(0.9,1)$. The resulting set is shown in \Cref{fig:bin_foliation_1}. The result of the algorithm is indeed a good approximation of the region of the red points, except for the white area between $x=0.2$ and $x=2.2$ in the upper part of the plots. To understand why the algorithm does not cover this area, let us zoom around the point $(0,0.5)$, looking at the foliation in $(-1,1) \times (-1,1)$, see \Cref{fig:foliation5e-6}. To separate the curves of the foliation more, we run the algorithm with $\delta = 5 \cdot 10^{-6}$. \Cref{al:SIMEXP2} is building the blue region as follows. First it builds the equivalence class -- the lowest light blue curve -- of $(0,0.5)$  -- the cyan dot in the picture -- using the SiMEC algorithm. Then it moves away from the point $(0,0.5)$ along the red curve, to change equivalence class, using SiMExp. Let us denote with $P_1$ this new point. At this point we repeat the procedure, obtaining the plot above. Once we arrive at the point in which the red curve hits the boundary $y=1$, we see that the right part equivalence class is convex and is lowering towards the bottom of the plot. To find the equivalence classes -- namely the curves of the foliation -- in the white area, we need to continue the procedure just described for a few points on the red curve in the region above $y=1$.
	
\begin{figure}[htbp]
		\centering
		\begin{tabular}{cc}
			\begin{tabular}{c}
				\begin{subfigure}{0.45\textwidth}
					\includegraphics[width=\textwidth]{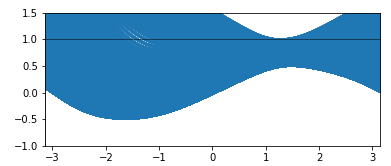}
					\caption{Reconstruction of the foliation of the input manifold near the starting point.}
					\label{fig:reconstruction_full}
				\end{subfigure}\\
				\begin{subfigure}{0.45\textwidth}
					\includegraphics[width=\textwidth]{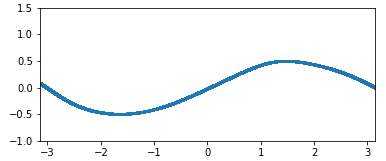}
					\caption{Reconstruction of the discriminating hypersurface of the classifier.}
					\label{fig:sin_separating}
				\end{subfigure}
			\end{tabular} &
			\begin{subfigure}{0.45\textwidth}
				\centering
				\includegraphics[width=1\textwidth]{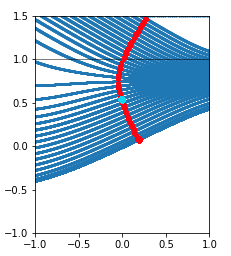}
				\caption{Zoom of the foliation around the starting point.}
				\label{fig:reconstruction_full_zoom}
			\end{subfigure}
		\end{tabular}
		\caption{Reconstruction of the foliation of the input manifold near the starting point. This time we allowed the algorithm to continue above $y=1$. Left panel: the entire foliation. Right panel: zoom of the foliation when we let the algorithm run over $y=1$. The SiMExp steps of \Cref{al:SIMEXP2} allow to recover the equivalence classes that are missing, by exploring the region outside the original domain but at the same time providing reliable reconstruction.}
	\end{figure}

	Indeed, running again \Cref{al:SIMEXP2} with $\delta=1\cdot 10^{-6}$ up to $y=1.5$ yields the set in \Cref{fig:reconstruction_full}. To space the curves of the foliation, we chose $\delta = 5 \cdot 10^{-6}$, see \Cref{fig:reconstruction_full_zoom}. Zooming around the point $(0,0.5)$ we can see that all the equivalence classes covering the white area are passing trough points of the red curve produced by SiMExp with $y>1$. Another way to solve this problem without continuing to run the the algorithm above $y=1$ would be try considering different starting point in the same equivalence class of $(0,0.5)$.
	As a final application, we build the separating hypersurface  -- in this case a line -- between red and blue points. To this end, we run \Cref{al:SIMEXP2} starting from a point which is red with a probability of approximatively $50 \%$. In this experiment we build the separating surface starting from $(0,0.-0.01499)$, whose output is $0.50382$ and we find all the points which are red with probability $[0.50382-\varepsilon,0.50382+\varepsilon]$ with $\varepsilon = 0.1$.
	The separating surface produced by the algorithm, see \Cref{fig:sin_separating}, is a curve approximating $y=\sin(x)$, as one expects.
	\begin{figure}[h!]
		\centering
		
	\end{figure}
	
	\section{Conclusions}
	In this work, starting from the geometric framework introduced in \cite{BeMa21}, we propose two algorithms: first we studied the SiMEC algorithm, which builds an approximation of the equivalence classes of neural network $\mathcal{N}$ -- solving the problem of finding the preimage of a point $p$ in the output manifold $M_n$ -- presenting both the underlying theory and some numerical experiments. Then, before tackling the case of the preimage of a real interval, we discussed how to explore the input manifold $M_0$ changing equivalence classes with the SiMExp algorithm. At last we studied how to build the preimage of an interval using both SiMEC and SiMExp, including an application to a binary classifier. We showed that once a neural network is properly trained, the equivalence classes are well reconstructed, as well as families of equivalence classes. We applied the proposed method to a thermodynamics problem and to a classification task, showing that one can easily reconstruct the separating surface.
	
	Future work will involve the extension  of the  SiMEC and SiMExp algorithms to the case of a neural network from $M_0 \subset \mathbb{R}^{m_0}$ to $M_n \subset \mathbb{R}^{m_n}$ for $m_0,m_n \in \mathbb{N}$. Moreover, further developments will be done for convolutional neural network and also frameworks that encompass also non smooth activation function.
	
	The possible applications of the presented approaches are vary: on one hand one can employ the SiMEC algorithm to explore the class of equivalence of a point $p\in M_n$, and this exploration actually means that \emph{new synthetic data} can be generated. On the other hand, the coupled role of the two prosed algorithms can provide useful insights on the behaviour of some class of neural networks, for example image classifiers that suffer from small perturbation on the input manifold, providing completely unreliable results. A further possible application is in medical imaging: for example, this exploration of the equivalence classes may give some indication on how much diagnostic images can be overexposed (or underexposed) without affecting the classification result.
	
	\bibliographystyle{elsarticle-num}
	\bibliography{biblio}
	
\end{document}